\documentclass[lettersize, journal]{IEEEtran}
\usepackage{amsmath, amsfonts}
\usepackage{algorithmic}
\usepackage{algorithm}
\usepackage{array}
\usepackage{textcomp}
\usepackage{stfloats}
\usepackage{url}
\usepackage{verbatim}
\usepackage{graphicx}
\usepackage{cite}
\usepackage{multirow}
\usepackage{booktabs}
\usepackage{subfigure}
\usepackage{diagbox}
\usepackage{bbding}
\DeclareUnicodeCharacter{2212}{\ensuremath{-}}
\usepackage{subcaption}

\begin{document}

\title{Dual-State Personalized Knowledge Tracing with Emotional Incorporation}
\author{Shanshan Wang, 
        Fangzheng Yuan, 
        Keyang Wang,
        Xun Yang,  
        Xingyi Zhang, 
        Meng Wang
\thanks{Shanshan Wang and Fangzheng Yuan are with the Information Materials and Intelligent Sensing Laboratory of Anhui Province, and Institute of Physical Science and Information Technology, Anhui University, Hefei 230601, China.(e-mail: wang.shanshan@ahu.edu.cn, q22301192@stu.ahu.edu.cn). }
\thanks{Keyang Wang is with the Zhejiang Dahua Technology Co., Ltd.(e-mail: 20181202010t@alu.cqu.edu.cn). }
\thanks{Xun Yang is with the Department of Electronic Engineering and Information Science, School of information Science and Technology, University of Science and Technology of China, Hefei, 230026, China (e-mail: xyang21@ustc.edu.cn).}
\thanks{Xingyi Zhang is with the Key Laboratory of intelligent computing and Signal Processing, Ministry of Education, and the school of Computer Science and Technology, Anhui University, Hefei, 230601, China (e-mail: xyzhanghust@gmail.com).}
\thanks{Meng wang is with the School of Computer Science and Information Engineering, Hefei University of Technology, Hefei, 230009, China (e-mail: wangmeng@hfut.edu.cn)}}



\maketitle

\begin{abstract}

Knowledge tracing has been widely used in online learning systems to guide the students' future learning.  
However, most existing KT models primarily focus on extracting abundant information from the question sets and explore the relationships between them, but ignore the personalized student behavioral information in the learning process.
This will limit the model's ability to accurately capture the personalized knowledge states of students and reasonably predict their performances.
To alleviate this limitation, we explicitly models the personalized learning process by incorporating the emotions, a representative personalized behavior in the learning process, into KT framework.
Specifically, we present a novel Dual-State Personalized Knowledge Tracing with Emotional Incorporation model to achieve this goal:
Firstly, we incorporate emotional information into the modeling process of knowledge state, resulting in the Knowledge State Boosting Module.  
Secondly, we design an Emotional State Tracing Module to monitor students' personalized emotional states, and propose an emotion prediction method based on personalized emotional states.
Finally, we apply the predicted emotions to enhance students' response prediction.
Furthermore, to extend the generalization capability of our model across different datasets, we design a transferred version of DEKT, named Transfer Learning-based Self-loop model (T-DEKT).
Extensive experiments show our method achieves the state-of-the-art performance.

\end{abstract}

\begin{figure}
	\centering  
	\subfigbottomskip=4pt 
	\subfigcapskip=-3pt 
	\subfigure[Emotional distribution and response accuracy under each distribution.]{
		\includegraphics[width=0.93\linewidth]{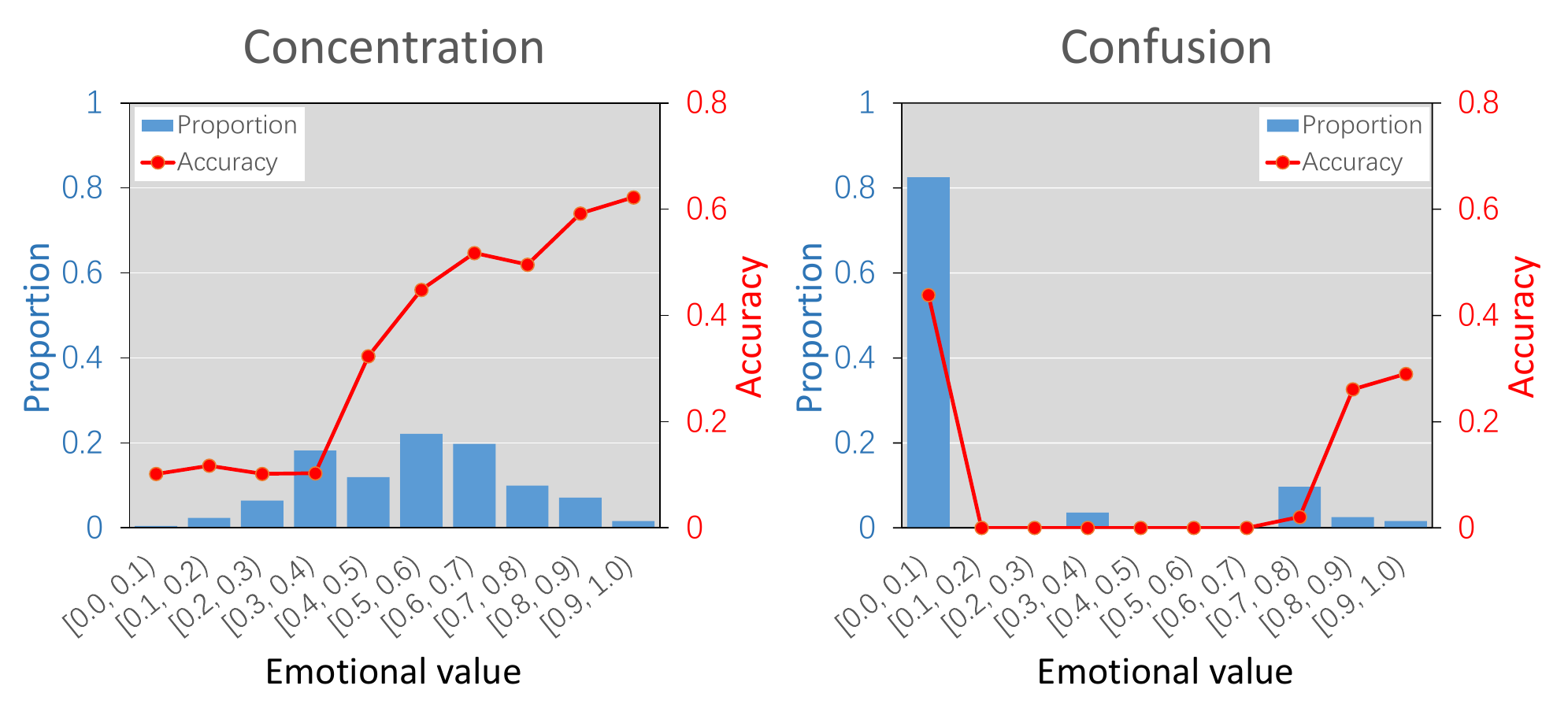}}
	  \\
	\subfigure[A toy example of students’ practice process and knowledge tracing.]{
		\includegraphics[width=0.93\linewidth]{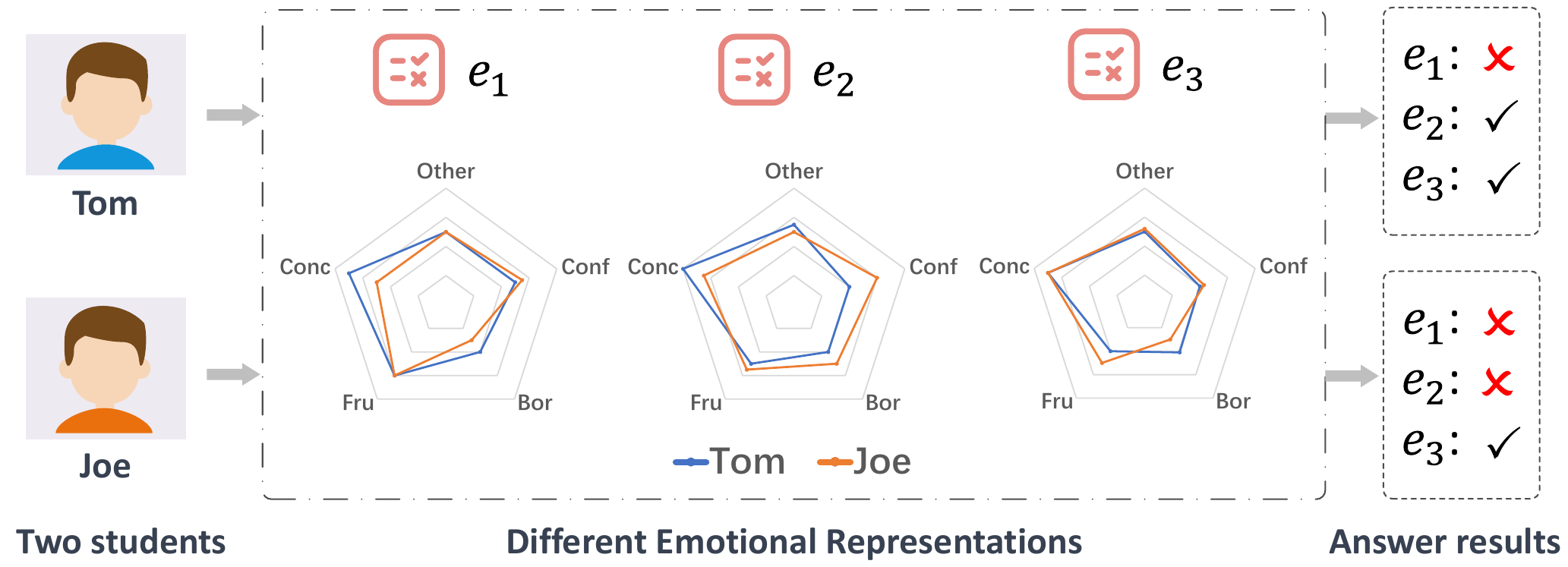}}
	\caption{(a) presents the data information of two emotional attributes in the ASSISTchall dataset.  
 The bar chart illustrates the distribution of emotions for each interval with a size of 0.1, while the line chart depicts the corresponding answer accuracy within each interval.
(b) presents the changes in four types of emotions during the completion of the same sequence of questions for two students with different inherent traits, Tom and Joe. The four emotions referred to as concentration, boredom, confusion and frustration, respectively. }
    \label{motivation}
\vspace{-0.3cm}
\end{figure}


\section{Introduction}
In recent years, the rapid development of intelligent education systems \cite{desmarais2012review, anderson2014engaging, abdelrahman2023knowledge} has played an indispensable role in achieving higher quality education.
Knowledge Tracing(KT), as an emerging research field, leverages machine learning and related technologies to dynamically monitor the evolving knowledge states of students using education-related data \cite{dowling2001automata, pardos2013adapting, wang2022neuralcd}.
The KT \cite{corbett1994knowledge, pelanek2017bayesian, wu2017modeling, shin2021saint+, zengzhen} task aims to capture changes in a student's knowledge state from past exercise interactions and predict future performance in solving questions. 
It plays a crucial role in real-time assessment of students' proficiency, enabling teachers to tailor personalized coaching plans for students in online learning environments \cite{harrell2013learner, SCD, liu2019exploiting}.
The KT methods can be divided into two categories: 
(1) Traditional methods. 
Such as Bayesian Knowledge Tracing (BKT) \cite{corbett1994knowledge, baker2008more, yudelson2013individualized} and Item Response Theory (IRT) \cite{van1997item, pardos2011kt} models. 
BKT models the student's knowledge state as a latent variable using a Hidden Markov Model.   
IRT models assess a student's ability by measuring both the student's proficiency and the difficulty level of the questions.
(2) Deep Learning-based methods.  
Deep Knowledge Tracing (DKT) \cite{piech2015deep} is the first to introduce neural networks into the knowledge tracing field and achieve remarkable results \cite{abdelrahman2019knowledge}.  
Exercise-aware Knowledge Tracing (EKT) \cite{liu2019ekt} incorporates \IEEEpubidadjcol text information for exercise to enhance knowledge tracing modeling.  
Difficulty Matching Knowledge Tracing (DIMKT) \cite{shen2022assessing} explores the impact of question difficulty information.

Despite the promising results shown by previous methods, there is still a significant limitation that most advanced KT models primarily focus on the attributes of the exercises themselves \cite{do2017integrating, dwivedi2018learning, wang2019adaptive}, but ignore the complex and personalized student behavioral information in the learning process. 
They assume students will always have the same knowledge increment after completing the same exercises and achieving the same results. 
However, in real-world scenarios, due to differences in students' intrinsic traits and cognitive attributes, they can generate personalized  behavioral information, especially emotional representation information, when answering the same questions. 
As a result, they always give quite inconsistent answers under the guidance of different emotions. 
We choose four highly learning-relevant emotions: concentration, frustration, boredom, and confusion, and visualize a set of representative exercise process in Figure \ref{motivation}(b), where two students answer the same questions with different emotions.
It can be clearly found that the student \textit{Joe} generates significantly more negative emotions (e.g., boredom and confusion) than student \textit{Tom} when facing the exercise $e_2$. Naturally, guided by these negative emotions, \textit{Joe} gives the wrong answer while \textit{Tom} does not.
It is evident from this visualization that emotions have a certain impact on a student's question-solving process and performance. 
Additionally, Figure \ref{motivation}(a) illustrates the distribution of two emotions in the ASSISTChall dataset, as well as the accuracy of answering questions in each distribution interval. We observe that the response accuracy under concentration distribution  follows a Z-score distribution, indicating a close relationship between emotions and response results.
Although later methods, such as Individualized Knowledge Tracing (IEKT) \cite{long2021tracing}, attempt to alleviate this limitation by modeling individual students' characteristics, they still rely solely on the raw information from the exercises themselves and do not involve student-specific information, fundamentally.
Therefore, leveraging student behavior for modeling KT remains a focal point worthy of attention.

In fact, we comprehensively analyze the students' interaction process and argue that the impact of personalized emotions can be categorized into two aspects: 
Firstly, the emotional information generated during a student's question-solving process can be considered as an indicator of learning quality.
A learning interaction that generates positive emotions is always  regarded as a high-quality learning process.
This helps students absorb and understand the learning content better after answering questions, making improvement in knowledge state more efficient.
Secondly, the emotional information can also influence the student's understanding of the question and the mobilization of their own knowledge state.
For example, negative emotions can lead to incomplete analysis of the question, indirectly causing students to focus on irrelevant information when recalling existing knowledge.
This can interfere with rational thinking about the problem, hinder the flexible application of knowledge, and ultimately result in a decrease in the accuracy of the response.

In order to address the above limitation and monitor the specific impacts of personalized emotions in the question-solving process, in this paper, we propose a Dual-State Personalized Knowledge Tracing with Emotional Incorporation (DEKT) model, which explicitly model emotions into the KT task from two aspects.
Firstly, we introduce a Knowledge State Boosting Module.      
Specifically, the core of this module is the  Emotion-Boosted learning gain.      
It generates personalized knowledge state for each student by exploring and simulating the impact mechanism of emotions during the process of knowledge accumulation.
Secondly, we attempt to introduce the emotional information into the response prediction to achieve more personalized and accurate performance prediction of exercises. 
However, we face a serious limitation: the student's actual emotions can not be utilized in the prediction phase before answering the questions in the standard KT task (emotional attributes contain response label information \cite{pardos2014affective,wang2015towards,pardos2013affective}). 
To address this limitation, we propose a method for emotional representation prediction based on personalized emotional state.
Specifically, similar to the knowledge state, we design an Emotional State Tracing Module to monitor the student's personalized emotional state at each learning interaction and further explicitly predict the student's emotions based on this personalized emotional state.
This module captures the inherent emotional traits of students from their past emotional patterns, and the predicted emotions do not contain response label information related to the current time.
Finally, we apply a Comprehensive Prediction Module that integrates the predicted personalized emotional representations into the response process to infer the student's performance in the next exercise.
Additionally, some publicly available datasets in the education domain do not provide any emotional information. 
Therefore, we further propose an extended version of DEKT based on transfer learning to make it applicable to more realistic educational scenarios.
The main contributions of our work can be summarized as follows:
\begin{itemize}
		\item[$\bullet$]This paper introduces a novel Dual-State Personalized Knowledge Tracing with Emotional Incorporation (DEKT) model.     
        To our knowledge, DEKT is the first model to explicitly explore and simulate the impact of emotions in the KT task.
		\item[$\bullet$]We incorporate the students' emotions, a representative personalized behavior in the learning process, into the modeling process of knowledge state and propose an Knowledge State Boosting Module, which aims to generate personalized knowledge state for each student by exploring and simulating the impact mechanism of emotions in the process of knowledge accumulation.
		\item[$\bullet$]We propose a method for emotional representation prediction based on personalized emotional state.    
        This method monitors students' personalized emotional states during each learning interaction using an Emotional State Tracing Module and accurately predicts students' personalized emotions based on the personalized emotional states during each exercise.
        \item[$\bullet$]We further propose an extended version of DEKT, called Transfer Learning-based Self-loop model (T-DEKT), based on transfer learning to enhance the generality of our method on datasets that do not provide any emotional information.  
        Extensive experiments on three public datasets show that our method can outperform the state-of-the-art KT models.
\end{itemize}

\section{RELATED WORKS}
\subsection{Knowledge Tracing}
With the widespread popularity of online education \cite{cully2019online,he2022multi}, the task of knowledge tracing has also rapidly advanced.
The KT model primarily consists of three categories, with the most classic being the probability models. Among these, the Bayesian Knowledge Tracing (BKT) \cite{corbett1994knowledge} model is the most prominent, which employs Hidden Markov Models to model students' knowledge states as latent variables. Another type of model is the logistic model, which uses the logistic function. Common examples of logistic models include Item Response Theory and Factor Analysis models.
With the advancement of artificial intelligence, researchers introduce deep learning models. One of the most classic models, Deep Knowledge Tracing (DKT) \cite{piech2015deep}, is the first to apply deep learning to KT models. DKT takes a student's historical interactions as input and uses RNN/LSTM \cite{hochreiter1997long} to simulate the student's knowledge state.
The Dynamic Key-Value Memory Network (DKVMN) \cite{zhang2017dynamic} and Deep Graph Memory Networks (DGMN) \cite{abdelrahman2022deep} introduce memory-enhanced neural networks.
To make more efficient use of original information, the Exercise-aware Knowledge Tracing (EKT) \cite{liu2019ekt} and Difficulty Matching Knowledge Tracing (DIMKT) \cite{shen2022assessing} enhance KT capabilities by utilizing text information and difficulty information to augment exercise representations.
Convolutional Knowledge Tracing (CKT) \cite{shen2020convolutional} introduces Convolutional Neural Networks (CNN) \cite{lecun2015deep}.
The Self-Attention Knowledge Tracing model (SAKT)  \cite{pandey2019self} is the first to apply Transformer models \cite{vaswani2017attention} to the field of knowledge tracing and introduces self-attention mechanisms to model long-term dependencies between learning interactions.
Subsequently, the Context-aware Attentive Knowledge Tracing (AKT) \cite{ghosh2020context} model combines attention-based models with interpretable model components inspired by psychological measurement models.
To achieve more accurate knowledge state modeling, the Learning Process-consistent Knowledge Tracing (LPKT) \cite{shen2021learning} model explicitly incorporates information related to both response time and the time interval between interactions.
Subsequently, LPKT-S \cite{shen2022monitoring}, IEKT \cite{long2021tracing}, and QIKT \cite{chen2023improving} attempt to model students using basic information.

\subsection{Emotional Effect}
In fact, the close relationship between learning and emotions is not a new concept. 
Despite the evident link between learning and emotions, research in this area is relatively scarce. 
It is only in recent years that researchers increasingly pay more attention to the role of emotions in the learning process \cite{pekrun2002academic}. 
These studies explore the impact of emotions on learning engagement, self-regulation, academic performance, and outcome evaluation \cite{efklides2001metacognitive, sansone2014don}.
Additionally, researchers also focus on the positive and negative nature of emotions and their influence on learning.
Researchers highlight the prospects and potential for understanding the complexity of the learning process through the study of emotions.  
Although it is clear that emotions are intimately intertwined with learning, the specific reasons why they influence the learning process are the subject of extensive experimental analysis and validation by researchers. 
One of the most prominent theories in this area is the ``Mood Congruence Hypothesis \cite{bower1981mood}", which suggests that emotional state can assist cognitive processes. 
Specifically, when in a positive emotion (e.g., feeling happy), individuals are more inclined to pay attention to positive information (such as successful exam experiences), while in a negative emotion (e.g., feeling sad), they tend to focus on negative information (such as exam failures). 
This consistency effect is attributed to the characteristics of our brain's structure, where information is organized through associations and semantic similarity. When similarity is strong and associations are tight, information activation is more likely to occur.
Subsequently, Schwarz \cite{schwarz1990feelings} propose the ``Emotion as Information Theory". 
He views emotions as additional information in the learning process, with positive emotions serving as a positive signal and negative emotions as a negative factor. 
These studies provide valuable insights into our deeper understanding of the importance of emotions in the field of education.

\section{PRELIMINARY}

\subsection{PROBLEM DEFINITION}
In the intelligent education system, suppose there is a group of students, denoted as $S=\{s_1, s_2, \ldots, s_I\}$. A set of exercises is denoted $X=\{e_1, e_2, \ldots, e_J\}$. A set of concepts is denoted $C=\{c_1, c_2, \ldots, c_M\}$.
Generally, Upon assigning exercises to students, they initiate their learning process. Throughout this journey, they allocate specific time to complete individual exercises, thereby constructing an ever-evolving knowledge state based on the information they have acquired. 
In this study, we introduce students' emotional information. we posit that not only their personalized knowledge state will change, but their personalized emotional state will also continue to fluctuate. Consequently, we extend the singular learning process into two parallel processes: One focuses on constructing personalized knowledge state during the learning process, while the other focuses on building personalized emotional state during the experiential process. The learning process can be represented as $LP=\{(e_{1}, at_{1}, a_{1}), {it}_{1}, \ldots, (e_{t}, at_{t}, a_{t}), {it}_{t}\} $, and the experiential process can be represented as $EP=\{(e_{1}, v_{1}), ..., (e_{t}, v_{t})\}$. In these processes, tuples $(e_t, at_{t}, a_t)$ represent the basic learning unit in the learning process, while tuples $(e_t, v_t)$ represent the basic emotional unit in the emotional experience process. 
$e_t$ represents the exercise representation that incorporates some properties of the exercise, $v_t$ represents  the primary emotional representation values exhibited by the student during answering questions, $at_t$ represents the time spent by the student in answering $e_t$, and $a_t$ represents the binary correctness label (1 for correct, 0 for incorrect).

\begin{figure}[t]
\centering
\includegraphics [width=\linewidth]{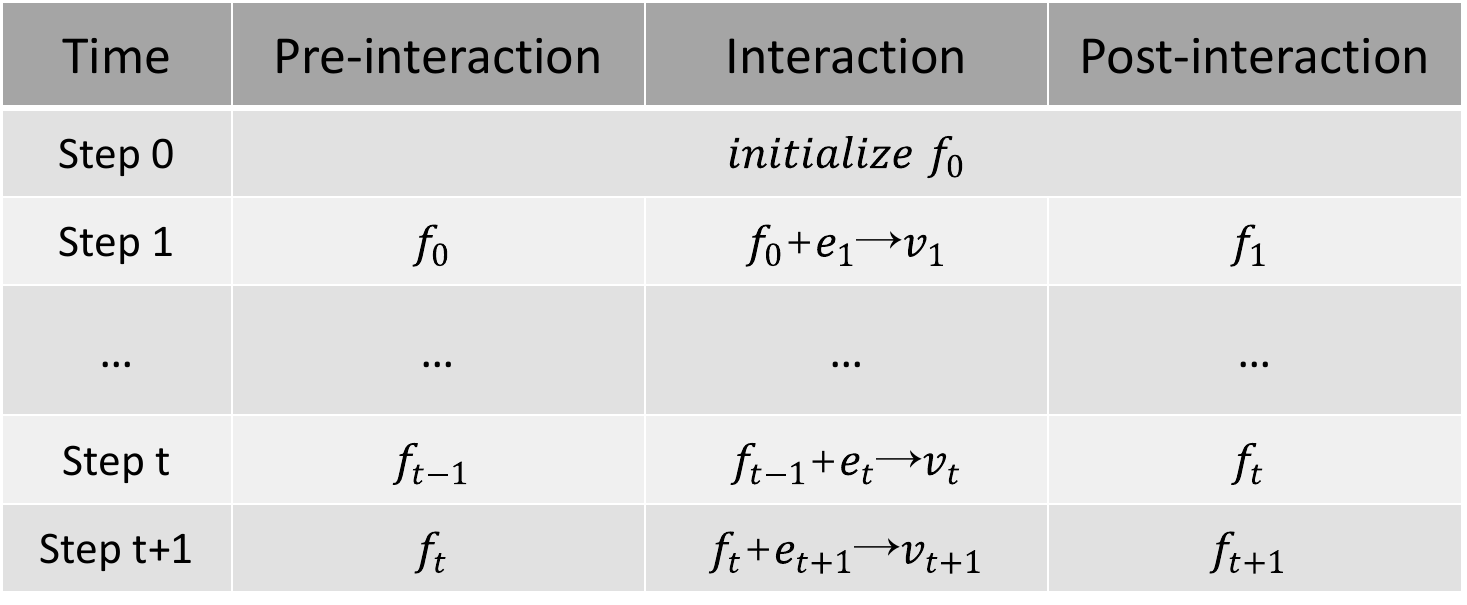}
\vspace{-0.2cm}
\caption{The changes in the personalized emotional state during a student's answering process.}
\label{Interpreting}
\vspace{-0.3cm}
\end{figure}

\section{Interpreting the Emotion Prediction Based on Personalized Emotional State}
\subsection{Emotional Representation and Emotional State }
Emotional representations refer to specific emotional experiences or outward expressions, such as frustration or confusion. 
They manifest as visible behavior. 
On the other hand, personalized emotional state represents individualized traits within students. 
it is an invisible state that can be expressed through specific emotional representation in particular situations.
Let’s start with a simple example: two students, \textit{A} and \textit{B}, answer the same exercise, $e_1$. 
Due to differences in their intrinsic emotional traits, they perceive $e_1$ differently. 
Student \textit{A} finds $e_1$ to be a challenging problem, while student \textit{B} sees it as less difficult.
Consequently, student \textit{A} experiences a sense of frustration, while student \textit{B} does not.
This can be represented mathematically as follows:
\begin{align}
StudentA&:&trait_A+e_1 & = Frustration\\StudentB&:&trait_B+e_1 & =No\enspace Frustration, 
\end{align}
here, $trait_A$ and $trait_B$ represent the inherent emotional traits of students \textit{A} and \textit{B}, respectively, while $Frustration$ and $No Frustration$ denote the specific emotions they generate.
From the above two formulas, the emotions generated during interaction are determined by the inherent traits at the student's emotional level.
In this paper, we define the inherent traits at the student's emotional level as the personalized emotional state, represented by $f$. 
The specific emotion values generated are defined as emotional representations, denoted by $v$.
Therefore, to predict the students' emotional representation when facing a question, it is essential not only to have necessary information about the question but also to understand the current personalized emotional state of the student.
\subsection{Tracing Principles of Personalized Emotional State }
The standard KT task can be viewed as a response prediction process, described as follow:
\begin{align}
h+e_{*} & = a_{*}, 
\end{align}
here, $h$ represents the student's personalized knowledge state, and $a$ denotes the response prediction result.
Similarly, our emotion prediction process runs parallel to the response prediction process, which can be defined as:
\begin{align}\label{F}
f+e_{*} & = v_{*}, 
\end{align}
here, $f$ represents  the student's personalized emotional state, and $v$ represents the generated emotional representation.
Hence, we represent the current KT task as the a combination of response prediction and emotion prediction tasks. 
Consequently, each interaction in the KT learning sequence is formalized as the fundamental unit $\{e_t, a_t, v_t\}$.
During the student's answering process, the changes in personalized emotional state are illustrated in the Figure \ref{Interpreting}.
It can be clearly observed that the personalized emotional state before and after each interaction is different. 
Therefore, we believe that students can generate an explicit change in personalized emotional state during each interaction.
Furthermore, students' personalized emotional states are always determined by their unique learning sequence.
Therefore, we can capture relationships from each interaction and utilize the fluctuations in personalized emotional state after each interaction to model the complete personalized emotional state of each student in the learning process. 
Specific modeling details will be provided in the methodology section.

\begin{figure*}[t]
\centering
\includegraphics [width=\linewidth]{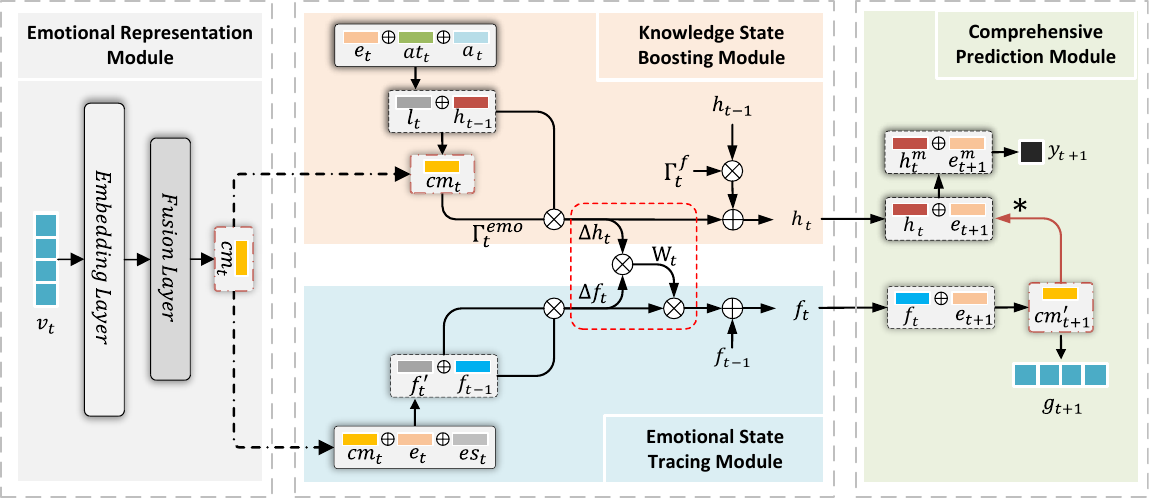}
\vspace{-0.2cm}
\caption{ 
The main structure of our DEKT model. We show the processing pipeline of DEKT at step $t$. In this timestep, we model two parallel processes. 
In the first process, the inputs are the exercise $e_t$, the answer time $at_t$, and the student's answer $a_t$, and the output is the knowledge state $h_t$ after answering. 
In the second process, the inputs are the exercise $e_t$, and the emotional embedding ${cm}_t$, and the output is the emotional state  $f_t$  after answering.
Moreover, we will predict the student’s performance $y_{t+1}$ at timestep $t+1$.
}
\label{DEKT}
\vspace{-0.2cm}
\end{figure*}

\section{METHODOLOGY}
In this chapter, we use LPKT as our basic framework to formally introduce our DEKT method. 
As illustrated in the Figure \ref{DEKT}, our DEKT model consists of four parts: 1) Emotional Representation Module, 2) Knowledge State Boosting Module, 3) Emotional State Tracing Module, and 4) Comprehensive Prediction Module. 
Subsequently, in order to enhance the generalizability of our model across datasets, especially for those without emotional attributes, we introduce an extended version of DEKT called Transfer Learning-based Self-loop model (T-DEKT).

\subsection{Emotional Representation Module}
As shown in Figure \ref{DEKT}, $v_t$ is an emotion vector composed of four real emotional representation values: concentration, boredom, confusion and frustration. 
We firstly apply an embedding layer to learn their respective embedded representations, denoted as $E_{t}^{conc}$, $E_{t}^{bor}$, $E_{t}^{conf}$, and $E_{t}^{fru}$. 
Next, given the varying impact and degrees of influence of these four emotions on the learning process, we employ a fusion layer to combine the embeddings of these four emotions into a comprehensive emotional representation embedding ${cm}_t$, which can harmonize their diversity. 
This approach ensures a unified treatment of distinct emotional factors.
\begin{align}
{cm}_t = W_1^T[E_{t}^{conc} \oplus E_{t}^{bor} \oplus E_{t}^{conf} \oplus E_{t}^{fru}]+b_1, 
\end{align}
where $\boldsymbol{W_1}\in\mathbb{R}^{(4d_k)\times d_k}$ is the weight matrix, $\boldsymbol {b_1}\in\mathbb{R}^{d_k}$ is the bias term.
Here, $E_{t}^{conc}$, $E_{t}^{bor}$, $ E_{t}^{conf}$ and $E_{t}^{fru}$ represent the embeddings of the four distinct emotions.

\subsection{Knowledge State Boosting Module}
In this section, we simplify the learning process of LPKT.   
Subsequently, we introduce emotional information into the learning process, resulting in Emotion-Boosted learning gain.    This aims to achieve more diverse and personalized knowledge acquisition after each interaction, ultimately leading to a more robust knowledge state.

\subsubsection{Simplified version of the learning process}
We simplify the process of modeling knowledge state of LPKT and propose a simplified version of the learning process.  
This allows our method to be based on LPKT but not be dependent on it, achieving a general-purpose model.  
During this process, We formalize each learning interaction as a basic learning unit $(e_{t}, at_{t}, a_{t})$.  
For extracting the learning embedding $l_t$ of each basic learning unit, we concatenate $e_t, at_t, a_t$ together and apply a perceptron to deeply fuse the exercise embedding, answer time embedding, and answer embedding as follows:
\begin{align}
l_t & = W_2^T[e_t\oplus at_t\oplus a_t]+b_2, 
\end{align}
where $\boldsymbol{W_2}\in\mathbb{R}^{(3d_k)\times d_k}$ is the weight matrix, $\boldsymbol {b_2}\in\mathbb{R}^{d_k}$ is the bias term.
Subsequently, we connect $l_t$ with the knowledge state $h_{t-1}$ from the previous time step .
This allows us to capture information from the current learning unit $l_t$ that was originally not present in $h_{t-1}$, thereby obtaining the knowledge gain after this interaction.
Specifically, to focus on the knowledge state relevant to the current exercise's knowledge concepts, we first multiply $h_{t-1}$ by the knowledge concept vector $q_{et}$ for the current exercise, resulting in the relevant knowledge state $\tilde{h}_{t-1}$.
\begin{align}
\tilde{h}_{t-1} & = q_{et}\cdot h_{t-1}, 
\end{align}
where $\cdot$ represents the element-wise multiplication between vectors. 
Then, the simplified version of learning gain modeling is defined as follows:
\begin{align}
lg_t & = tanh(W_3^T[l_t\oplus\tilde{h}_{t-1}]+b_3), 
\end{align}
where $\boldsymbol{W_3}\in\mathbb{R}^{(2d_k)\times d_k}$ is the weight matrix, $\boldsymbol {b_3}\in\mathbb{R}^{d_k}$ is the bias term.
The modeling of the learning gate that controls a student's capacity to absorb knowledge is defined as follows:
\begin{align}\label{H}
\Gamma_{t}^{lg} & = \sigma(W_{4}^{T}[l_{t} \oplus\tilde{h}_{t-1}]+b_{4}), 
\end{align}
where $\boldsymbol{W_4}\in\mathbb{R}^{(2d_k)\times d_k}$ is the weight matrix, $\boldsymbol {b_4}\in\mathbb{R}^{d_k}$ is the bias term.
Then, multiplying $lg_t$ and $\Gamma_{t}^{lg}$ yields the actual learning gains for the t-th interaction, and similarly, to focus on relevant concepts, we multiply $\Delta h_t$ by $q_{et}$ to obtain the relevant knowledge gains.
\begin{align}
\Delta h_t & = \Gamma_{t}^{lg}\cdot((lg_{t}+1)/2) \\
\widetilde {\Delta h_t} & = \boldsymbol{q}_{e_{t}}\cdot \Delta h_t.
\end{align}
To ensure that knowledge is acquired in each interaction, we map the output range of the $tanh$ function to [0, 1] through a linear transformation, thus ensuring that $\Delta h_t$ is always positive.
\subsubsection{Emotion-Boosted learning gain}
Although we can capture the actual knowledge gains $\Delta h_t $ after each learning interaction by associating learning embedding $l_t$ with the student's relevant knowledge state $\tilde{h}_{t-1}$, this process relies heavily on exercise-related information and is less influenced by personalized student behavior.
Therefore, the learning gains obtained in this way always  lack personalization and diversity.
So we design the emotion-boosted learning gains by introducing the personalized  emotional representations of students into the learning process.
For instance, even if two students with equal abilities both answer the same question correctly, their learning gains may differ due to the varying emotions they experience during the answering process.

Specifically, we still use the learning unit and the relevant knowledge state from the previous time step to obtain the knowledge gains $lg_{t}$. 
However, when calculating the learning gate $\Gamma_{t}^{lg}$, we incorporate the influence of emotions. Positive emotions indicate a high-quality learning interaction. 
Therefore, we connect the emotional embedding ${cm}_t$ and the learning unit $l_t$ from the t-th interaction and combine them with the relevant knowledge state $\tilde{h}_{t-1}$ from the previous time step $t-1$. 
Then we apply a perceptron to generate a emotion-boosted learning gains gate $\Gamma_{t}^{emo}$, as: 
\begin{align}
\Gamma_{t}^{emo} & = \sigma(W_{5}^{T}[ {{cm}_t}\oplus l_{t} \oplus\tilde{h}_{t-1}]+b_{5}), 
\end{align}
where $\boldsymbol{W_5}\in\mathbb{R}^{(3d_k)\times d_k}$ is the weight matrix, $\boldsymbol {b_5}\in\mathbb{R}^{d_k}$ is the bias term, ${cm}_t$ represents the comprehensive emotional embedding during the answering process at time $t$.
After obtaining the emotion-boosted learning gains gate $\Gamma_{t}^{emo}$, we replace the standard learning gate $\Gamma_{t}^{lg}$ in Eq.~\eqref{H} with it to generate a more personalized actual knowledge gains $\Delta h_t$ that is controlled by personalized emotions of student.
\subsubsection{Forgetting Module}
Subsequently, we adopt LPKT's modeling of forgetting.
\begin{align}
\Gamma_{t}^{f} & = \sigma(W_{6}^{T}[h_{t-1}\oplus \Delta h_t\oplus it_{t}])+b_{6}), 
\end{align}
where $\boldsymbol{W_6}\in\mathbb{R}^{(3d_k)\times d_k}$ is the weight matrix, $\boldsymbol {b_6}\in\mathbb{R}^{d_k}$ is the bias term, $\Gamma_{t}^{f}$ is the forget gate.
After modeling the learning and forgetting processes, we obtain the personalized knowledge state $h_t$ after the t-th interaction as follows:
\begin{align}
h_{t} & = \widetilde {\Delta h_t} +\Gamma_{t}^{f}\cdot h_{t-1}.
\end{align}
The introduction of emotions imbues the knowledge state $h_t$ with personalized emotional information about students.

\subsection{Emotional State Tracing Module}
The primary goals of our emotional representation prediction are to simulate and trace the personalized emotional states of students at each moment.
Given that a student's personalized emotional state is dynamically evolving, our focus is on capturing this change.
In fact, in every interaction process, we can explicitly construct two general pieces of emotional state information: 1) the fundamental personalized emotional state before current interaction. 2) the new personalized emotional gains caused by current interaction.  
Taking step $t$ in Figure \ref{Interpreting} as an example, $f_{t-1}$ represents the fundamental personalized emotional state, and the difference between $f_t$ and $f_{t-1}$ represents the new personalized emotional gains, which is defined as $\Delta f_t$ in this study.  
Through these two emotional state pieces of information, we can calculate the personalized emotional state $f_t$ of the student after the t-th interaction.

\subsubsection{Capture New Personalized Emotional Gains}
We begin the emotional state tracing process by connecting the emotional embedding ${cm}_t$ and the exercise representation $e_t$.  
Then, we employ a fully connected layer with a $sigmoid$ activation to capture the temporary personalized emotional state $f^{'}_t$ of the student during the current interaction.
At the same time, considering the mutual influence between  the exercisers' emotions and the interaction process, we further incorporate two emotion-sensitive interaction factors, i.e., answer result $a_t$ and answer time $at_t$, into the temporary personalized emotional state $f^{'}_t$. Specifically, we design a weighting mechanism as shown in Eq.~\eqref{es} to integrate these two factors into an emotion-sensitive embedding $es_t$.
\begin{align}\label{es}
es_t & = \Bigg(Softmax\Bigg(\begin{bmatrix}{{cm}_t}\\1\end{bmatrix}\times\begin{bmatrix}at_t\\a_t\end{bmatrix}\Bigg)\Bigg).\begin{bmatrix}at_t\\a_t\end{bmatrix}, 
\end{align}
where, $\mathrm{Softmax}(u_i) = e^{u_i}/\sum_je^{u_j}$.
Therefore, the formula for calculating the current temporary personalized emotional state $f^{'}_t$ is as follow:
\begin{align}
f_t^{^{\prime}} & = \sigma(W_7^T[e_t\oplus {cm}_t\oplus es_t])+b_7, 
\end{align}
where $\boldsymbol{W_7}\in\mathbb{R}^{(3d_k)\times d_k}$ is the weight matrix, $\boldsymbol {b_7}\in\mathbb{R}^{d_k}$ is the bias term.

Subsequently, we eliminate the fundamental personalized emotional state information $f_{t-1}$ from the temporary personalized emotional state $f^{'}_t$, resulting in the new personalized emotional gains $\Delta f_t$ present in the current interaction.
Specifically, we first obtain the personalized emotional gains $AEC_t$ through a fully connected network with a $sigmoid$ activation function.
Afterwards, Considering that not all information can be retained, we also design an emotion gate $\Gamma_t^{AEC}$ to control emotional information transmission:

\begin{align}
AEC_t  & = tanh(W_8^T[f_t^{^{\prime}}\oplus f_{t-1}])+b_8\\
\Gamma_t^{AEC}  & = \sigma(W_{9}^T[f_t^{^{\prime}}\oplus f_{t-1}])+b_{9}, 
\end{align}
where $\boldsymbol{W_8}\in\mathbb{R}^{(2d_k)\times d_k}$ and $\boldsymbol{W_9}\in\mathbb{R}^{(2d_k)\times d_k}$  are the weight matrices, $\boldsymbol {b_8}\in\mathbb{R}^{d_k}$ and $\boldsymbol {b_9}\in\mathbb{R}^{d_k}$ are the bias terms.
Finally, we multiply $AEC_t$ by $\Gamma_t^{AEC}$ to obtain the ultimate new emotional gains $\Delta f_t$ during the current interaction process.
\begin{align}
\Delta f_t & = AEC_t\cdot\Gamma_t^{AEC}.
\end{align}

\subsubsection{Update Personalized Emotional State}
Later, we persist the new emotional gains $\Delta f_t$ into the fundamental emotional state $f_{t-1}$ to obtain the final personalized emotional state $f_t$ after the t-th interaction.
However, the calculated personalized emotional gains $\Delta f_t$ from the previous section may include some temporary information that can not be persistently updated into the student's personalized emotional state.
For example, the emotional information generated by guessing and slipping behaviors while students answer questions has obvious transience. 
Preserving such information may lead to erroneous influences on students.
At the same time, through this learning interaction, students can identify the irrational aspects of their fundamental emotional states.
For instance, if students grasp the key to solving a certain concept during this learning session, then the negative emotions they previously held toward this concept will become irrational, and thus should be discarded.
Therefore, we utilize the personalized knowledge gains $\widetilde{\Delta h_t}$ and the personalized emotional gains $\Delta f_t$ obtained from the current interaction to balance the updating ratio between new emotional gains and the fundamental emotional state.
Specifically, we multiply the personalized knowledge gains $\widetilde{\Delta h_t}$ by the new personalized emotional gains $\Delta f_t$ to obtain the updating weight $W_t$. 
The formula is as follows:
\begin{align}
W_t & = Softmax(W_{10}^T[\Delta f_t\cdot\widetilde{\Delta h_t}])+b_{10}, 
\end{align}
where $\boldsymbol{W_{10}}\in\mathbb{R}^{d_k\times d_k}$ is the weight matrix, $\boldsymbol {b_{10}}\in\mathbb{R}^{d_k}$ is the bias term.
Next, we use the weight $W_t$ to calculate the weighted sum of $f_{t-1}$  and $\Delta f_t$, obtaining the personalized emotional state $f_t$ after the t-th interaction. 
The formula is as follows:
\begin{align}
f_t & = W_t\cdot\Delta f_t+(1-W_t)\cdot f_{t-1}.
\end{align}
This indicates that after this interaction, erroneous information is removed from the fundamental personalized emotional state $f_{t-1}$, while new personalized emotional gains information $\Delta f_t$ is added.

\subsection{Comprehensive Predicting Module}
Now that we have obtained the student's personalized knowledge state $h_t$ and personalized emotional state $f_t$, we first apply  $f_t$ to complete the emotion prediction at time $t+1$, and then use the predicted emotional representation to enhance response prediction.
\subsubsection{Emotion Prediction Based on Personalized Emotional State}
After simulating the student's personalized emotional state $f_t$, we can easily predict the comprehensive emotional representation embedding ${cm}_{t+1}^{'}$ of student in the (t+1)-th interaction.
Specifically, following the formula \eqref{F}, we concatenate the personalized emotional state $f_t$ with the exercise representation $e_{t+1}$ and apply a fully connected layer with a $sigmoid$ activation function to predict the possible emotional representation embedding ${cm}_{t+1}^{'}$ that the student may exhibit while solving the next exercise:
\begin{align}
{cm}_{t+1}^{'} & = \sigma(W_{11}^T[f_t\oplus e_{t+1}])+b_{11}, 
\end{align}
where $\boldsymbol{W_{11}}\in\mathbb{R}^{(2d_k)\times d_k}$ is the weight matrix, $\boldsymbol {b_{11}}\in\mathbb{R}^{d_k}$ is the bias term.

\subsubsection{Emotion-Boosted Response Prediction}
In a real learning environment, given $e_{t+1}$ to the students, they will generate the personalized exercise understanding $e^m_{t+1}$ and personalized mobilization of their knowledge state $h^m_t$ under the guidance of their personalized emotional embedding embedding ${cm}_{t+1}^{'}$, The formulas are as follows:
\begin{align}
e_{t+1}^m& = {cm}_{t+1}^{'}\cdot e_{t+1}\\
h_t^m& = {cm}_{t+1}^{'}\cdot h_t.
\end{align}
Therefore, in the next interaction, we use the personalized exercise understanding and personalized mobilization of their knowledge state to infer the student's performance $y_{t+1}$ on $e_{t+1}$. 
We first concatenate $e^m_{t+1}$ with $h^m_t$, then project them to the output layer by a fully connected network with a $sigmoid$ activation:
\begin{align}
y_{t+1} & = \sigma(W_{12}^T[e_{t+1}^m\oplus h_{t}^m]+b_{12}), 
\end{align}
where $\boldsymbol{W_{12}}\in\mathbb{R}^{(2d_k)\times d_k}$ is the weight matrix, $\boldsymbol {b_{12}}\in\mathbb{R}^{d_k}$ is the bias term.
Finally, we decompose the predicted comprehensive emotional representation embedding ${cm}_{t+1}^{'}$ into the corresponding four specific emotional representation values $g_{t+1}$, which are then used to calculate the loss against the actual emotion values $v_{t+1}$. 
Here, $v_{t+1}$ represents the ground truth of the emotional representation values generated at time $t+1$.
Specifically, we utilize a fully connected layer to transform this 1x128 emotional embedding into a 1x4 emotion vector. 
Each item in this vector corresponds to the predicted value of a specific emotion.
\begin{align}
g_{t+1} & = \sigma(W_{13}^T[{cm}_{t+1}^{'}]+b_{13}), 
\end{align}
where $\boldsymbol{W_{13}}\in\mathbb{R}^{d_k\times d_k}$ is the weight matrix, $\boldsymbol {b_{13}}\in\mathbb{R}^{d_k}$ is the bias term.

\begin{figure}[t]
\centering
\includegraphics [width=\linewidth]{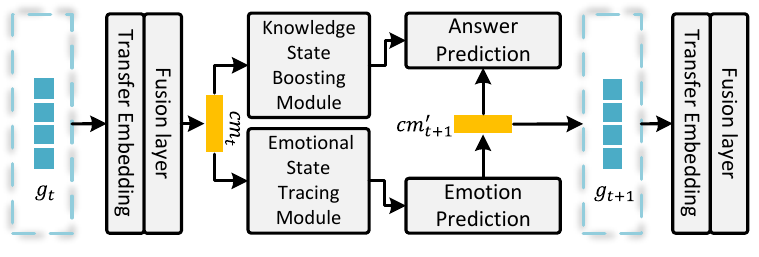}
\vspace{-0.2cm}
\caption{The architecture of T-DEKT model}
\label{T-DEKT}
\vspace{-0.2cm}
\end{figure}


\subsection{Objective Function}
For response prediction, we choose the cross-entropy log loss between the predicted probability $y_t$ and the actual answer $a_t$ as the objective function.
\begin{align}
\mathcal{L}_{\mathrm{1}} = -\sum_{t = 1}^T(a_t\log y_t+(1-a_t)\log(1-y_t)).
\end{align}
For emotion prediction, the objective function involves minimizing the mean squared error between the predicted emotion value $g$ and the true emotion value $v$. 
\begin{align}
\mathcal{L}_{\mathrm{2}}  = -\sum_{t= 1}^{T} (g_{t} - v_{t})^2, 
\end{align}
where $T$ represents the time step, $g$ represents the predicted emotion value and $v$ represents the ground truth of the emotion value.
The joint learning of parameters is achieved by minimizing the combined loss of response prediction and emotion prediction.
\begin{align}
\mathcal{L} & = \mathcal{L}_{\mathrm{1}} +\mathcal{L}_{\mathrm{2}}.
\end{align}

\subsection{T-DEKT model} 
Due to the difficulty in collecting emotional information in the past, most datasets in the field of education do not provide emotional attributes.  
In such cases, the Knowledge State Boosting Module in our DEKT method lacks a source of information.  
To address this limitation and ensure the applicability of our method to the majority of datasets, we propose an improved version of DEKT called Transfer Learning-based Self-loop model (T-DEKT), as illustrated in the Figure \ref{T-DEKT}.
Specifically, we first randomly initialize the four emotional representation values at time $t_0$ as the initial input $g_0$ for the model. 
Subsequently, we use the model to predict the emotions $g_1$ at time $t_1$, and then use $g_1$ as the input for the next time step to predict $g_2$, and so on until the end of the sequence.
We believe that even across different datasets, similar emotions have similar effects on learning.
At the same time, to enhance the constraint on emotion prediction in the model, we do not learn the embeddings for the four emotions from scratch.  
Instead, we transfer these embeddings from models trained on other datasets with emotion supervision information.
This approach starts from the intrinsic emotional representations and explores the commonalities of emotional representations cross different datasets instead of relearning new emotion embeddings, which can better fit our autoregressive model.

\section{EXPERIMENTS}
In this chapter, we introduce the real-world datasets used in our study and conduct experiments.  
We show baseline methods, compare experimental results for student response prediction, and show DEKT’s interpretability from diverse angles.
\subsection{Datasets} 
We utilize three real-world datasets to assess the effectiveness of DEKT and T-DEKT.  
Table \ref{dataset} presents the statistical data for all datasets.  
Below, we provide a brief description of each dataset.

\begin{itemize}
		\item[$\bullet$] \textbf{ASSISTments 2012\footnote{https://sites.google.com/site/assistmentsdata/home/2012-13-school-data-with affect}(ASSIST2012)} is collected from ASSISTments \cite{feng2009addressing}, an online tutoring system created in 2004. It includes data from the 2012-2013 academic year for emotion prediction. Records with no concepts and questions were removed. Additionally, ASSIST2012 includes four emotion-related attributes."
  
        \item[$\bullet$] \textbf{ASSISTments Challenge\footnote{https://sites.google.com/view/assistmentsdatamining/dataset}(ASSISTChall)}  is a dataset from the 2017 ASSISTments Data Mining Competition. This dataset records the usage of the ASSISTments blended learning platform by middle school students from 2004 to 2007. In this dataset, the student's learning sequences are significantly longer than those in ASSIST2012. Similar to ASSIST2012, this dataset provides seven emotion-related attributes.

		\item[$\bullet$] \textbf{EdNet-KT1\footnote{http://ednet-leaderboard.s3-website-ap-northeast-1.amazonaws.com/}}  is a dataset collected over two years, encompassing all student interactions with the system by Santa. EdNet \cite{choi2020ednet} offers four datasets with varying levels of abstraction. In this paper, we employ the simplest form, EdNet-KT1, composed of student question-solving logs. If an exercise involves multiple knowledge concepts, we only consider the first knowledge concept. Unfortunately, this dataset does not provide emotion-related attributes.

\end{itemize}

\begin{table}[]
\resizebox{1\linewidth}{!}{
\renewcommand{\arraystretch}{1.2}
\begin{tabular}{c|ccc}
\hline
\multirow{2}{*}{Statistics} & \multicolumn{3}{c}{Datasets}                                                  \\ \cline{2-4} 
                            & \multicolumn{1}{c}{ASSIST2012} & \multicolumn{1}{c}{ASSISTchall} & EdNet-KT1 \\ \hline
Students                    & \multicolumn{1}{c}{29018}      & \multicolumn{1}{c}{1709}        & 784309    \\ 
Exercises                   & \multicolumn{1}{c}{53091}      & \multicolumn{1}{c}{3162}        & 12375     \\ 
Concepts                    & \multicolumn{1}{c}{265}        & \multicolumn{1}{c}{102}         & 141       \\ 
Answer Time                 & \multicolumn{1}{c}{26747}      & \multicolumn{1}{c}{1326}        & 2292      \\ 
Interval Time               & \multicolumn{1}{c}{29748}      & \multicolumn{1}{c}{2839}        & 13143     \\ 
Avg.length                  & \multicolumn{1}{c}{93.45}      & \multicolumn{1}{c}{551.68}      & 121.48    \\ 
\hline
\end{tabular}
}
\caption{Statistics of all datasets}
\vspace{-0.15cm}
\label{dataset}
\vspace{-0.15cm}
\end{table}

\begin{table*}[h]
\resizebox{1\linewidth}{!}{
\renewcommand{\arraystretch}{1.2}
\begin{tabular}{@{}c|cccc|cccc|cccc@{}}
\toprule
\multirow{2}{*}{Method} & \multicolumn{4}{c|}{ASSIST2012}                                                                                   & \multicolumn{4}{c|}{ASSISTchall}                                                                                  & \multicolumn{4}{c}{EdNet-KT1}                                                                                     \\ \cmidrule(l){2-13} 
                        & \multicolumn{1}{c|}{RMSE}   & \multicolumn{1}{c|}{AUC}    & \multicolumn{1}{c|}{ACC}    & \multicolumn{1}{c|}{$r^2$} & \multicolumn{1}{c|}{RMSE}   & \multicolumn{1}{c|}{AUC}    & \multicolumn{1}{c|}{ACC}    & \multicolumn{1}{c|}{$r^2$} & \multicolumn{1}{c|}{RMSE}   & \multicolumn{1}{c|}{AUC}     & \multicolumn{1}{c|}{ACC}    & \multicolumn{1}{c}{$r^2$} \\ \midrule
DKT \cite{piech2015deep}                      & \multicolumn{1}{l|}{0.4241} & \multicolumn{1}{l|}{0.7289} & \multicolumn{1}{l|}{0.7360} & 0.1468                  & \multicolumn{1}{l|}{0.4471} & \multicolumn{1}{l|}{0.7213} & \multicolumn{1}{l|}{0.6907} & 0.1425                  & \multicolumn{1}{l|}{0.4508} & \multicolumn{1}{l|}{0.6836}  & \multicolumn{1}{l|}{0.6889} & 0.1008                 \\
DKVMN \cite{zhang2017dynamic}                 & \multicolumn{1}{l|}{0.4261} & \multicolumn{1}{l|}{0.7228} & \multicolumn{1}{l|}{0.7329} & 0.1398                  & \multicolumn{1}{l|}{0.4498} & \multicolumn{1}{l|}{0.7110} & \multicolumn{1}{l|}{0.6852} & 0.1320                  & \multicolumn{1}{l|}{0.4538} & \multicolumn{1}{l|}{0.6741}  & \multicolumn{1}{l|}{0.6843} & 0.0913                \\
SAKT  \cite{pandey2019self}                    & \multicolumn{1}{l|}{0.4258} & \multicolumn{1}{l|}{0.7233} & \multicolumn{1}{l|}{0.7339} & 0.1403                   & \multicolumn{1}{l|}{0.4626} & \multicolumn{1}{l|}{0.6605} & \multicolumn{1}{l|}{0.6694} & 0.0822                  & \multicolumn{1}{l|}{0.4524} & \multicolumn{1}{l|}{0.6794}  & \multicolumn{1}{l|}{0.6862} & 0.0964                 \\
DTransformer \cite{yin2023tracing}            & \multicolumn{1}{l|}{0.4118} & \multicolumn{1}{l|}{0.7698} & \multicolumn{1}{l|}{0.7509} & 0.2004                  & \multicolumn{1}{l|}{0.4371} & \multicolumn{1}{l|}{0.7506} & \multicolumn{1}{l|}{0.7078} & 0.1791                  & \multicolumn{1}{l|}{0.4291} & \multicolumn{1}{l|}{0.7553}  & \multicolumn{1}{l|}{0.7089} & 0.1837                 \\

EKT  \cite{liu2019ekt}                       & \multicolumn{1}{l|}{0.4132} & \multicolumn{1}{l|}{0.7597} & \multicolumn{1}{l|}{0.7473} & 0.1497                  & \multicolumn{1}{l|}{0.4389} & \multicolumn{1}{l|}{0.7438} & \multicolumn{1}{l|}{0.6985} & 0.1667                 & \multicolumn{1}{l|}{0.4356} & \multicolumn{1}{l|}{0.7328}  & \multicolumn{1}{l|}{0.7114} & 0.1625              \\ 
AKT  \cite{ghosh2020context}                    & \multicolumn{1}{l|}{0.4100} & \multicolumn{1}{l|}{0.7707} & \multicolumn{1}{l|}{0.7535} & 0.2004                  & \multicolumn{1}{l|}{0.4364} & \multicolumn{1}{l|}{0.7501} & \multicolumn{1}{l|}{0.7080} & 0.1801                  & \multicolumn{1}{l|}{0.4297} & \multicolumn{1}{l|}{0.7557}  & \multicolumn{1}{l|}{0.7204} & 0.1842                 \\
IEKT \cite{long2021tracing}                  
& \multicolumn{1}{l|}{0.4105} & \multicolumn{1}{l|}{0.7618} & \multicolumn{1}{l|}{0.7506} & 0.2106               
& \multicolumn{1}{l|}{0.4228} & \multicolumn{1}{l|}{0.7812} & \multicolumn{1}{l|}{0.7155} & 0.2156                 
& \multicolumn{1}{l|}{0.4384} & \multicolumn{1}{l|}{0.7405}  & \multicolumn{1}{l|}{0.6945} & 0.1660                \\

LPKT \cite{shen2021learning}                 & \multicolumn{1}{l|}{0.4103} & \multicolumn{1}{l|}{0.7741} & \multicolumn{1}{l|}{0.7531} & 0.2094                 & \multicolumn{1}{l|}{0.4182} & \multicolumn{1}{l|}{0.7982} & \multicolumn{1}{l|}{0.7372} & 0.2549                  & \multicolumn{1}{l|}{0.4287} & \multicolumn{1}{l|}{0.7655} & \multicolumn{1}{l|}{0.7220} & 0.1867                \\ 

LPKT-S  \cite{shen2022monitoring}
& \multicolumn{1}{l|}{0.4065} & \multicolumn{1}{l|}{0.7803} & \multicolumn{1}{l|}{0.7564} & 0.2195              
& \multicolumn{1}{l|}{0.4175} & \multicolumn{1}{l|}{0.7991} & \multicolumn{1}{l|}{0.7421} & 0.2561             
& \multicolumn{1}{l|}{0.4259} & \multicolumn{1}{l|}{0.7668}  & \multicolumn{1}{l|}{0.7249} & 0.1935             \\

DIMKT  \cite{shen2022assessing}                & \multicolumn{1}{l|}{0.4059} & \multicolumn{1}{l|}{0.7806} & \multicolumn{1}{l|}{0.7563} & 0.2188                  & \multicolumn{1}{l|}{\underline{0.4156}} & \multicolumn{1}{l|}{\underline{0.8059}} & \multicolumn{1}{l|}{\underline{0.7426}} & \underline{0.2628}   	      & \multicolumn{1}{l|}{0.4238} & \multicolumn{1}{l|}{0.7721} & \multicolumn{1}{l|}{0.7254} & 0.1977                  \\ 

QIKT \cite{chen2023improving}
& \multicolumn{1}{l|}{\underline{0.4047}} & \multicolumn{1}{l|}{\underline{0.7814}} & \multicolumn{1}{l|}{\underline{0.7578}} & \underline{0.2210}             
& \multicolumn{1}{l|}{0.4166} & \multicolumn{1}{l|}{0.8040} & \multicolumn{1}{l|}{0.7411} & 0.2612             
& \multicolumn{1}{l|}{\underline{0.4218}} & \multicolumn{1}{l|}{\underline{0.7746}}  & \multicolumn{1}{l|}{\underline{0.7261}} & \underline{0.2012}          \\ \midrule \midrule
T-DEKT              & \multicolumn{1}{l|}{0.4032} & \multicolumn{1}{l|}{0.7831} & \multicolumn{1}{l|}{0.7594} & 0.2232                  & \multicolumn{1}{l|}{0.4103} & \multicolumn{1}{l|}{0.8155} & \multicolumn{1}{l|}{0.7567} & 0.2809                  & \multicolumn{1}{l|}{\textbf{0.4153}} & \multicolumn{1}{l|}{\textbf{0.7883}}  & \multicolumn{1}{l|}{\textbf{0.7362}} & \textbf{0.2373}                 \\ 
DEKT                   & \multicolumn{1}{l|}{\textbf{0.3988}} & \multicolumn{1}{l|}{\textbf{0.7923}} & \multicolumn{1}{l|}{\textbf{0.7680}} & \textbf{0.2353}                  & \multicolumn{1}{l|}{\textbf{0.3949}} & \multicolumn{1}{l|}{\textbf{0.8413}} & \multicolumn{1}{l|}{\textbf{0.7694}} &\textbf{0.3351}                  & \multicolumn{1}{c|}{$-$} & \multicolumn{1}{c|}{$-$}  & \multicolumn{1}{c|}{$-$} & $-$                 \\ \bottomrule \bottomrule
\end{tabular}
}

\caption{The results of various comparison methods in the student response prediction task are shown, when the input dimension of emotion embedding is set to 1000.  The best results are bold, "-" denotes no corresponding result, and "\_" represents the state-of-the-art existing results.}
\label{result}
\vspace{-0.3cm}
\end{table*}

\begin{table*}[t]
\vspace{0.5cm}
\resizebox{1\linewidth}{!}{
\begin{tabular}{@{}l|ccccc|cccc@{}}
\toprule
Methods            & Embedding  & Gain       & Expression & Exercise   & Interaction & RMSE   & AUC    & ACC    & $r^2$  \\ \midrule
DEKT-NoEmbedding   &            & \Checkmark & \Checkmark & \Checkmark & \Checkmark  & 0.3985 & 0.8342 & 0.7643 & 0.3226 \\
DEKT-NoGain        & \Checkmark &            & \Checkmark & \Checkmark & \Checkmark  & 0.4040 & 0.8252 & 0.7578 & 0.3039 \\
DEKT-NoExpression  & \Checkmark & \Checkmark &            & \Checkmark & \Checkmark  & 0.3995 & 0.8339 & 0.7639 & 0.3197 \\
DEKT-NoExercise    & \Checkmark & \Checkmark & \Checkmark &            & \Checkmark  & 0.3989 & 0.8344 & 0.7640 & 0.3216 \\
DEKT-NoInteraction & \Checkmark & \Checkmark & \Checkmark & \Checkmark &             & 0.3961 & 0.8400 & 0.7666 & 0.3311 \\ \midrule
DEKT               & \Checkmark & \Checkmark & \Checkmark & \Checkmark & \Checkmark  & 0.3949 & 0.8413 & 0.7694 & 0.3351 \\ \bottomrule
\end{tabular}
}
\caption{Ablation experiment results on ASSISTChall.}
\label{ablation}
\end{table*}

\begin{figure*}
	\centering  
	\subfigbottomskip=3pt 
	\subfigcapskip=1pt 
	\subfigure[Question Answering Sequence.]{
		\includegraphics[width=0.55\linewidth]{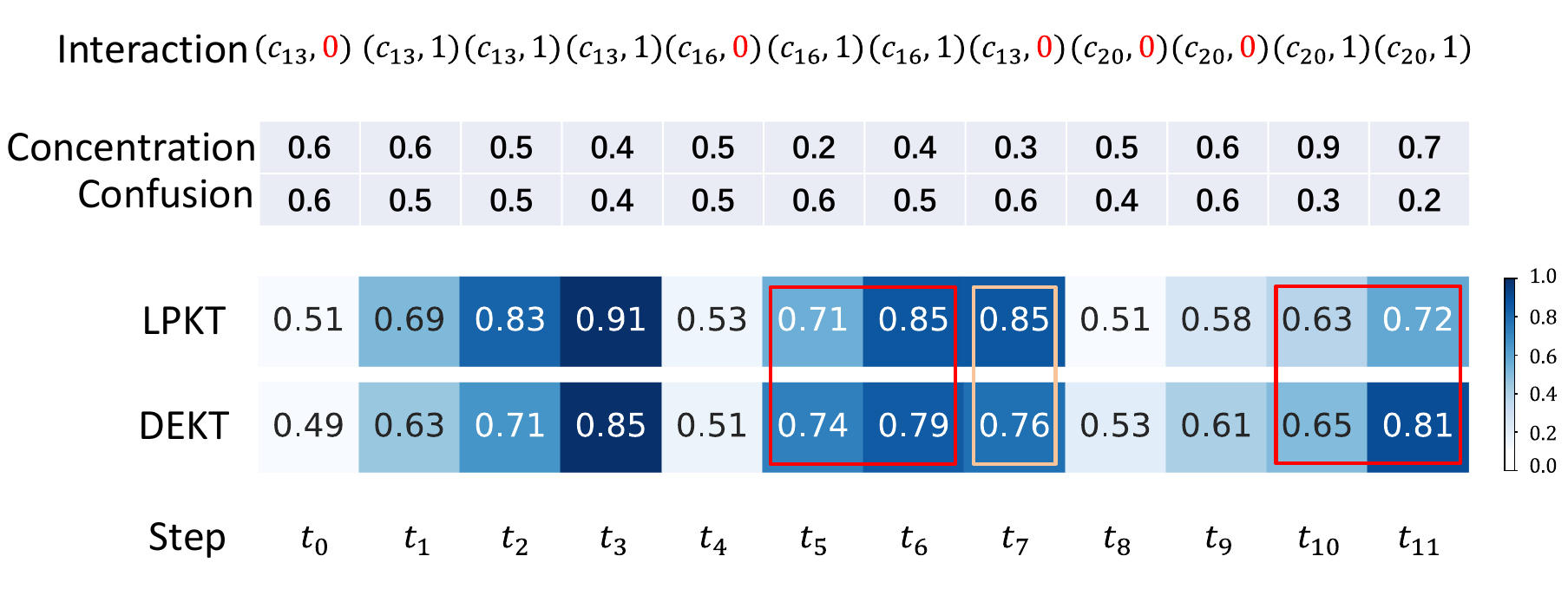}}       
	\subfigure[t-SNE Visualization.]{
		\includegraphics[width=0.43\linewidth]{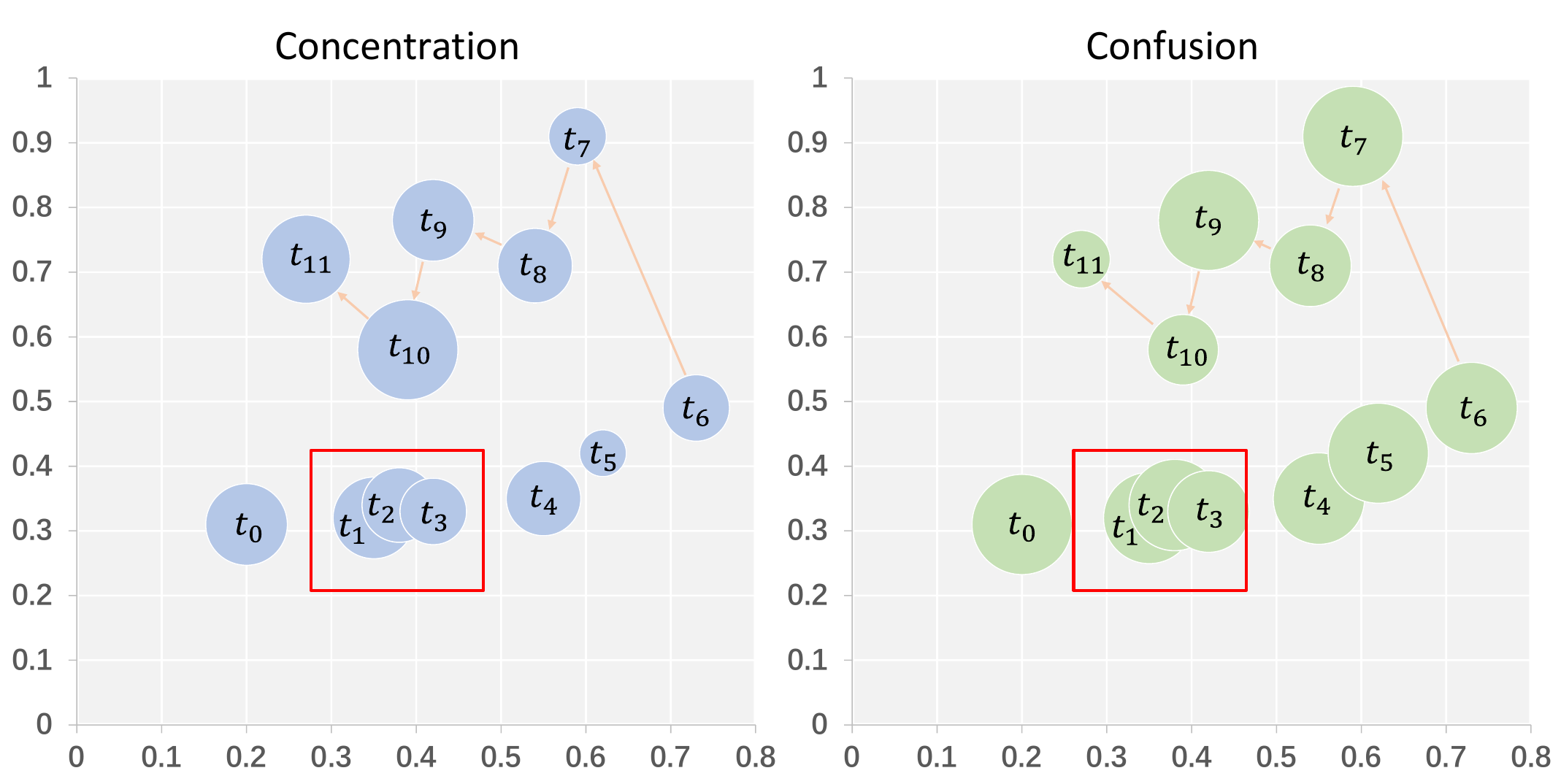}}     
	\caption{Illustrates the dynamic changes in the knowledge state $h_t$ and emotional state $f_t$ of a student throughout the process of consecutively solving 13 exercises  involving 3 different concepts.}
    \label{Visualization}
\end{figure*}

\subsection{Baseline Methods} 
We compare DEKT with previous methods. 
To ensure fairness, all baseline methods have been fine-tuned to achieve their best performance. 
All methods are implemented using PyTorch and trained on a Linux server. A brief introduction to all the baselines is provided below:
\begin{itemize}
		\item[$\bullet$] \textbf{DKT\cite{piech2015deep}} is the first method that utilizes Recurrent Neural Networks (RNN), specifically Long Short-Term Memory (LSTM), to model student's knowledge state.

		\item[$\bullet$] \textbf{DKVMN\cite{zhang2017dynamic}} introduces a key-value memory network that uses a static key matrix and a dynamic value matrix to store knowledge concepts and their corresponding concept states, respectively.

		\item[$\bullet$] \textbf{SAKT\cite{pandey2019self}} is the first model to incorporate attention mechanisms, weighting the importance of questions in the previous answering sequence. 

        \item[$\bullet$] \textbf{DTransformer\cite{yin2023tracing}} establishes an architecture from the problem level to the knowledge level, explicitly diagnosing learners' proficiency based on the mastery of each problem. It utilizes contrastive learning to maintain the stability of diagnosing knowledge state.

        \item[$\bullet$] \textbf{EKT\cite{liu2019ekt}} incorporates textual information of the questions into KT, including difficulty information.
        \item[$\bullet$] \textbf{AKT\cite{ghosh2020context}} is a context-aware attention-based knowledge tracing model. It utilizes two self-attention encoders to learn context-aware representations for both exercises and responses.
        \item[$\bullet$] \textbf{IEKT \cite{long2021tracing}} models the estimation of student cognition and knowledge acquisition.
        \item[$\bullet$] \textbf{LPKT\cite{shen2021learning}} utilizes two pieces of information, the learner's response time and the interval time, to calculate the learning gains and forgetting between two consecutive time steps. This helps model the learner's learning process.
        \item[$\bullet$] \textbf{LPKT-S \cite{shen2022monitoring}} further explicitly distinguishes the individual progress rate of each student based on LPKT.
        \item[$\bullet$] \textbf{DIMKT\cite{shen2022assessing}} utilizes difficulty information of both questions and concepts to model personalized cognition and personalized learning estimation for students.
        \item[$\bullet$] \textbf{QIKT \cite{chen2023improving}} learns question-centric cognitive representations and proposes an interpretable prediction layer based on Item Response Theory (IRT).
            
\end{itemize}

\subsection{Training Details} 
\subsubsection{Preprocessing}
First, we process the emotional attributes we intend to use. We discretize the original continuous emotion values ranging from 0 to 1. We discretize the continuous emotion values into 1000 categories by setting a discrete interval of 0.001, which corresponds to setting the input dimension of the Embedding layer to 1000.
Next, we sort all student learning records based on timestamps. We then set all input sequences to a fixed length, determined by the average sequence length of the dataset. Specifically, we set the fixed lengths of the ASSIST2012, ASSISTchall, and EdNet-KT1 datasets to 100, 500, and 150, respectively. 
For sequences that exceed the fixed length, we truncate them to match the fixed length. For sequences shorter than the fixed length, we pad them with zero vectors to reach the fixed length.

\subsubsection{Transfer training}
When training T-DEKT, we transfer the model parameters obtained from DEKT, including the embedding matrices for the four emotions and some of the weights (W) and biases (B) from the fully connected layers. We selectively freeze these parameters and continue training on the new dataset.

\subsubsection{Training setting}
We initialize the parameters of our models in the uniform distribution.    For all datasets, we perform standard 5-fold cross-validation for all models in which 80\% of students are selected as the training set (80\%) and validation set (20\%), and the rest 20\% were used as the testing set.
The input dimension of the Embedding layer is the number of categories with discrete emotion values, and the $d_k$ of the hidden layer is set to 128.  In order to prevent overfitting, we have set a dropout layer with a dropout rate of 0.2. 
The learning rate is 0.002.
 
\section{RESULTS AND DISCUSSION}
In this section, we will present our experimental results and discuss the outcomes of these experiments.

\subsection{Student Response Prediction}
We conduct comprehensive experiments to compare our DEKT and T-DEKT with all baseline methods on three datasets. 
To ensure the reliability of the evaluation results, all experiments are assessed using metrics such as Root Mean Square Error (RMSE), Area Under the Curve (AUC), Accuracy (ACC), and Pearson Correlation Coefficient Square ($r^2$). 
During the T-DEKT experiments on ASSIST2012 and ASSISTchall, we exclude the emotional attributes provided by the dataset.
Table \ref{result} presents the experimental results, corresponding to the average results of 5-fold cross-validation on the testing set.
we observe that both our DEKT and T-DEKT outperform all baseline methods.
Furthermore, we notice that on the ASSISTchall dataset, DEKT outperforms the state-of-the-art DIMKT model (with an AUC improvement of 4.39\%) and the baseline LPKT model (with an AUC improvement of 5.43\%), indicating that DEKT is more capable of accurately constructing emotional mechanisms in long sequence historical learning.
Although T-DEKT's improvement is somewhat less than DEKT, it still significantly outperforms the best baseline model. 
Particularly on the EdNet-KT1 dataset without emotional attributes, T-DEKT surpasses the state-of-the-art QIKT (with an AUC improvement of 1.77\%) and the baseline LPKT (with an AUC improvement of 2.98\%).
This indicates that our emotion prediction method is effective.
The performance gap between T-DEKT and DEKT is mainly due to the difference in the source of emotional information for the Knowledge State Boosting Module.
T-DEKT uses predicted emotions, while DEKT employs true emotions.
Our future work aims to enhance the accuracy of predicted emotions to minimize the performance gap between the two models.
The significant improvement of DEKT clearly demonstrates the advantage of incorporating emotional information to enhance the knowledge gains, emphasizing the positive correlation between more personalized knowledge state modeling and predictive performance.

\subsection{Ablation Experiments }
In this chapter, we conduct ablation experiments to clarify the roles of individual modules within the DEKT framework. We enhance five variations of DEKT by removing specific modules from the original DEKT model :
\begin{itemize}
		\item[$\bullet$] DEKT-NoEmbedding: This variant removes the embedding layer and fusion layer from the model and directly replaces the comprehensive emotional embedding with a four-dimensional emotion vector.
		\item[$\bullet$] DEKT-NoGain: This variant eliminates the influence of emotion on knowledge gain.
        \item[$\bullet$] DEKT-NoExpression: This variant eliminates the influence of emotion on the mobilization of knowledge state.
		\item[$\bullet$] DEKT-NoExercise: This variant removes the influence of emotion on exercise understanding.
        \item[$\bullet$] DEKT-NoInteraction: This variant removes the influence of knowledge gains on the calculation of weighting parameters when updating personalized emotional state.
\end{itemize}
We evaluate the impact of the above variant models on the final prediction performance, and present some noteworthy findings in Tabel \ref{ablation}.
Firstly, DEKT-NoGain exhibits the most significant performance decline. This suggests that incorporating emotional information in each interaction to acquire more personalized learning gains is meaningful, leading to a more accurate final knowledge state.
Subsequently, DEKT-NoEmbedding performs poorly, indicating that discretizing and embedding fusion of emotional values can help uncover deeper information within emotions.  
Additionally, both personalized exercise understanding and personalized mobilization of knowledge state contribute to the performance of DEKT.  However, the removal of personalized exercise understanding results in an even greater decrease, as it better reflects the extent to which students are influenced by emotions.
DEKT-NoInteraction also exhibits a certain degree of decline, suggesting that modeling emotional state is indeed closely related to modeling knowledge state.


\begin{table*}[h]
\resizebox{1\linewidth}{!}{
\begin{tabular}{@{}c|ccccc|ccccc@{}}
\toprule
Dataset & \multicolumn{5}{c|}{ASSISTchall}                                                                                 & \multicolumn{5}{c}{ASSIST2012}                                                                                  \\ \midrule
Emotion & \multicolumn{1}{c|}{Boredom}  & \multicolumn{1}{c|}{Concentration} & \multicolumn{1}{c|}{Confusion}& \multicolumn{1}{c|}{Frustration} & Mean & \multicolumn{1}{c|}{Boredom}  & \multicolumn{1}{c|}{Concentration} & \multicolumn{1}{c|}{Confusion}& \multicolumn{1}{c|}{Frustration} & Mean \\ \midrule
DEKT    & \multicolumn{1}{c|}{0.1069} & \multicolumn{1}{c|}{0.1624}        & \multicolumn{1}{c|}{0.2441}    & \multicolumn{1}{c|}{0.2323} & \underline{\textbf{0.1864}}   & \multicolumn{1}{c|}{0.0961} & \multicolumn{1}{c|}{0.1581}        & \multicolumn{1}{c|}{0.1194}   & \multicolumn{1}{c|}{0.1194}          &  \underline{\textbf{0.1232}}   \\
T-DEKT & \multicolumn{1}{c|}{0.1390} & \multicolumn{1}{c|}{0.1906}        & \multicolumn{1}{c|}{0.3735}    & \multicolumn{1}{c|}{0.3720}   & \underline{\textbf{0.2687}}     & \multicolumn{1}{c|}{0.1541} & \multicolumn{1}{c|}{0.3817}        & \multicolumn{1}{c|}{0.2410}  & \multicolumn{1}{c|}{0.3531} & \underline{\textbf{0.2824}}     \\ \bottomrule
\end{tabular}
-}
\caption{The performance of the emotion prediction on two datasets with emotion supervision information.  
RMSE (Root Mean Square Error) is adopted as the metric to measure the difference between the predicted emotions and the actual emotions.}
\label{emotions predicted table}
\end{table*}

\begin{figure*}
	\centering  
	\subfigbottomskip=2pt 
	\subfigcapskip=-5pt 
	\subfigure[Distribution of Emotion values predicted by DEKT (supervised).]{
		\includegraphics[width=0.93\linewidth]{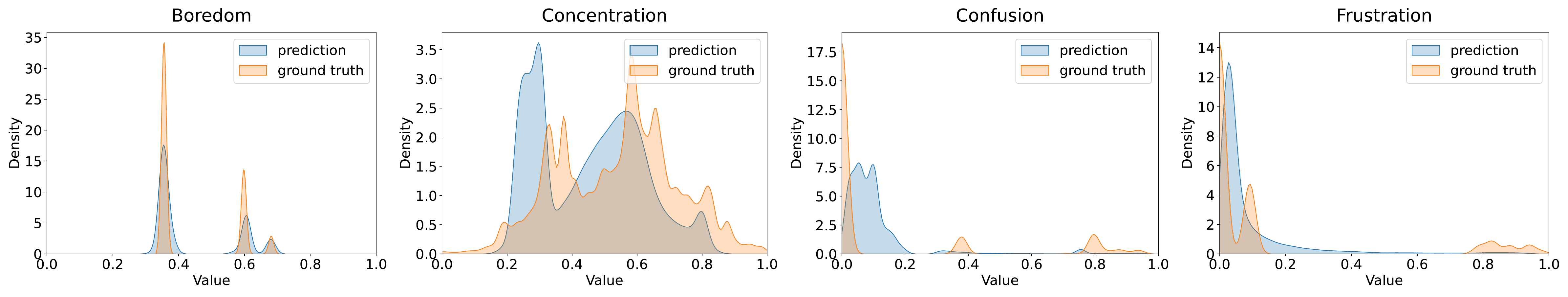}}
	  \\
	\subfigure[Distribution of Emotion values predicted by T-DEKT (unsupervised).]{
		\includegraphics[width=0.93\linewidth]{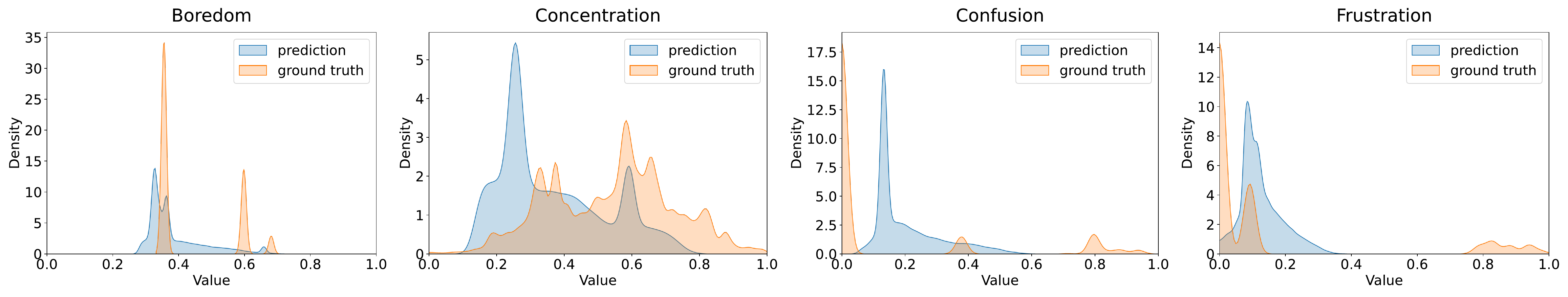}}
	\caption{Illustrates the comparison between predicted emotion distribution and actual emotion distribution under the two methods. The term "prediction" represents the predicted emotion, while "ground truth" denotes the actual emotion.}
    \label{emotions predicted figure}
\end{figure*}

\subsection{Analysis the Performance of Emotion Prediction}
In this section, we conduct experiments to demonstrate the performance of emotional representation prediction based on personalized emotional state.
The performance of our emotional representation prediction critically determines the generalization of our overall model on real-world datasets.
We calculate the root mean square error (RMSE) of DEKT and T-DEKT predictions for four emotions on the ASSIST2012 and ASSISTchall datasets.
As shown in Table \ref{emotions predicted table}, DEKT employs supervised emotion prediction, while T-DEKT utilizes unsupervised emotion prediction, relying entirely on the model's fitting process.
RMSE is computed between the true emotions and predicted emotions.
For DEKT, We achieve the average RMSE values of 0.1864 and 0.1232 respectively on two datasets, which clearly indicate that our DEKT can accurately predict personalized emotions of each exerciser according to the proposed emotional prediction method. For T-DEKT, although its performance is slightly poorer than DEKT, we still achieve the average RMSE values of 0.2687 and 0.2824 respectively.
In the absence of true emotional label constraint, T-DEKT achieves prediction performance comparable to DEKT solely relying on constraint from response outcomes and model fitting.
This not only demonstrates the superiority of our overall model structure but also validates the rationality of the emotion prediction method we designed.
Additionally, we observe that T-DEKT exhibits lower performance in predicting certain emotions, which is correlated with the distribution of true emotions.
We conduct experiments on the ASSISTchall dataset using both DEKT and T-DEKT, recording the specific predicted emotion values in both cases.
We then compare their distributions to their true emotion values, as shown in the Figure \ref{emotions predicted figure}.
We found that the emotion distributions predicted by T-DEKT closely resemble a standard normal distribution based on Z-scores.
This is because in the absence of true emotional constraint, emotion value prediction relies on the relationship between emotional level and response accuracy. 
Through model fitting, it is discovered that excessively high or low emotions lead to significant fluctuations in response accuracy. 
Therefore, under the constraint of response outcomes, the probability that occurs mediocre emotion values increases, while the probability that occurs excessively high or low emotion values decreases.
Let us take the ``Frustration” distribution under T-DEKT as an example, it would be a challenge for our emotion prediction method to perform well without true emotion values constraint in the extreme situations where true emotion values are concentrated near 0.

\begin{table*}[h]
\resizebox{1\linewidth}{!}{
\begin{tabular}{@{}c|cccc|cccc|cccc@{}}
\toprule
\multirow{2}{*}{Model} & \multicolumn{4}{c|}{ASSIST2012}                                                                                                      & \multicolumn{4}{c|}{ASSISTchall}                                                                                                     & \multicolumn{4}{c}{EdNet-KT1}                                                                                                        \\ \cmidrule(l){2-13} 
                       & \multicolumn{1}{c|}{RMSE}            & \multicolumn{1}{c|}{AUC}             & \multicolumn{1}{c|}{ACC}             & $r^2$               & \multicolumn{1}{c|}{RMSE}            & \multicolumn{1}{c|}{AUC}             & \multicolumn{1}{c|}{ACC}             & $r^2$               & \multicolumn{1}{c|}{RMSE}            & \multicolumn{1}{c|}{AUC}             & \multicolumn{1}{c|}{ACC}             & $r^2$              \\ \midrule
DKVMN                & \multicolumn{1}{l|}{0.4261} & \multicolumn{1}{l|}{0.7228} & \multicolumn{1}{l|}{0.7329} & 0.1398                  & \multicolumn{1}{l|}{0.4498} & \multicolumn{1}{l|}{0.7110} & \multicolumn{1}{l|}{0.6852} & 0.1320                  & \multicolumn{1}{l|}{0.4538} & \multicolumn{1}{l|}{0.6741}  & \multicolumn{1}{l|}{0.6843} & 0.0913                \\
DKVMN + DEKT            & \multicolumn{1}{c|}{\textbf{0.4090}} & \multicolumn{1}{c|}{\textbf{0.7416}} & \multicolumn{1}{c|}{\textbf{0.7480}} & \textbf{0.1603} & \multicolumn{1}{c|}{\textbf{0.4301}} & \multicolumn{1}{c|}{\textbf{0.7457}} & \multicolumn{1}{c|}{\textbf{0.7077}} & \textbf{0.2082} & \multicolumn{1}{c|}{-}               & \multicolumn{1}{c|}{-}               & \multicolumn{1}{c|}{-}               & -               \\
DKVMN + T-DEKT          & \multicolumn{1}{c|}{0.4164}          & \multicolumn{1}{c|}{0.7341}          & \multicolumn{1}{c|}{0.7422}          & 0.1521          & \multicolumn{1}{c|}{0.4426}          & \multicolumn{1}{c|}{0.7256}          & \multicolumn{1}{c|}{0.6943}          & 0.1830          & \multicolumn{1}{c|}{\textbf{0.4316}} & \multicolumn{1}{c|}{\textbf{0.6959}} & \multicolumn{1}{c|}{\textbf{0.7140}} & \textbf{0.1167} \\ \midrule \midrule
LPKT-S                   & \multicolumn{1}{l|}{0.4065} & \multicolumn{1}{l|}{0.7803} & \multicolumn{1}{l|}{0.7584} & {0.2195}              
& \multicolumn{1}{l|}{0.4175} & \multicolumn{1}{l|}{0.7991} & \multicolumn{1}{l|}{0.7421} & 0.2561             
& \multicolumn{1}{l|}{0.4259} & \multicolumn{1}{l|}{0.7668}  & \multicolumn{1}{l|}{0.7249} & 0.1935  \\
LPKT-S + DEKT             & \multicolumn{1}{c|}{\textbf{0.3984}} & \multicolumn{1}{c|}{\textbf{0.7963}} & \multicolumn{1}{c|}{\textbf{0.7698}} & \textbf{0.2401} & \multicolumn{1}{c|}{\textbf{0.3928}} & \multicolumn{1}{c|}{\textbf{0.8452}} & \multicolumn{1}{c|}{\textbf{0.7705}} & \textbf{0.3362} & \multicolumn{1}{c|}{-}               & \multicolumn{1}{c|}{-}               & \multicolumn{1}{c|}{-}               & -               \\ 
LPKT-S + T-DEKT           & \multicolumn{1}{c|}{0.4021}          & \multicolumn{1}{c|}{0.7842}          & \multicolumn{1}{c|}{0.7601}          & 0.2237          & \multicolumn{1}{c|}{0.3998}          & \multicolumn{1}{c|}{0.8207}          & \multicolumn{1}{c|}{0.7582}          & 0.2839          & \multicolumn{1}{c|}{\textbf{0.4132}} & \multicolumn{1}{c|}{\textbf{0.7894}} & \multicolumn{1}{c|}{\textbf{0.7382}} & \textbf{0.2383} \\ \midrule \midrule  
DIMKT                  & \multicolumn{1}{c|}{0.4059}          & \multicolumn{1}{c|}{0.7806}          & \multicolumn{1}{c|}{0.7563}          & 0.2188          & \multicolumn{1}{c|}{ 0.4156}          & \multicolumn{1}{c|}{ 0.8059}          & \multicolumn{1}{c|}{0.7426}          & 0.2628          & \multicolumn{1}{c|}{0.4238}          & \multicolumn{1}{c|}{ 0.7721}          & \multicolumn{1}{c|}{0.7254}          & 0.1977          \\
DIMKT + DEKT             & \multicolumn{1}{c|}{\textbf{0.3962}} & \multicolumn{1}{c|}{\textbf{0.7975}} & \multicolumn{1}{c|}{\textbf{0.7624}} & \textbf{0.2311} & \multicolumn{1}{c|}{\textbf{0.3971}} & \multicolumn{1}{c|}{\textbf{0.8326}} & \multicolumn{1}{c|}{\textbf{0.7668}} & \textbf{0.2981} & \multicolumn{1}{c|}{-}               & \multicolumn{1}{c|}{-}               & \multicolumn{1}{c|}{-}               & -               \\
DIMKT + T-DEKT          & \multicolumn{1}{c|}{0.4010}          & \multicolumn{1}{c|}{0.7888}          & \multicolumn{1}{c|}{0.7604}          & 0.2256          & \multicolumn{1}{c|}{0.4081}          & \multicolumn{1}{c|}{0.8123}          & \multicolumn{1}{c|}{0.7519}          & 0.2798          & \multicolumn{1}{c|}{\textbf{0.4122}} & \multicolumn{1}{c|}{\textbf{0.7880}} & \multicolumn{1}{c|}{\textbf{0.7352}} & \textbf{0.2216} \\ \midrule \bottomrule 
\end{tabular}
}
\caption{The results of using three classical process-based models to replace the baseline of our method to validate its generality.}
\label{comparison method}
\end{table*}

\subsection{Experiment on Method Generalization}
To further validate the effectiveness and generality of the method, we replace the baselines of DEKT and T-DEKT with other classical process-based KT models.  
Here, we select DKVMN, LPKT-S, and DIMKT.  
During the replacement process, some minor adjustments are necessary.  
We test the performance of these models on all three datasets.  
The experimental results are shown in Table \ref{comparison method}.  
we observe that in most cases, introducing our method leads to these different baselines can achieve significant performance improvements on all three datasets. 
Additionally, in our experiments, we do not pursue the best hyperparameters because it is not our focus.  
Furthermore, for non-process-based KT models such as DKT and SAKT, they are not very compatible with our method, because these models can not extract information from emotional data containing label information, and they lack the learning process like LPKT.  
We will address this issue in future work.

\begin{figure}[t]
\centering
\includegraphics [width=\linewidth]{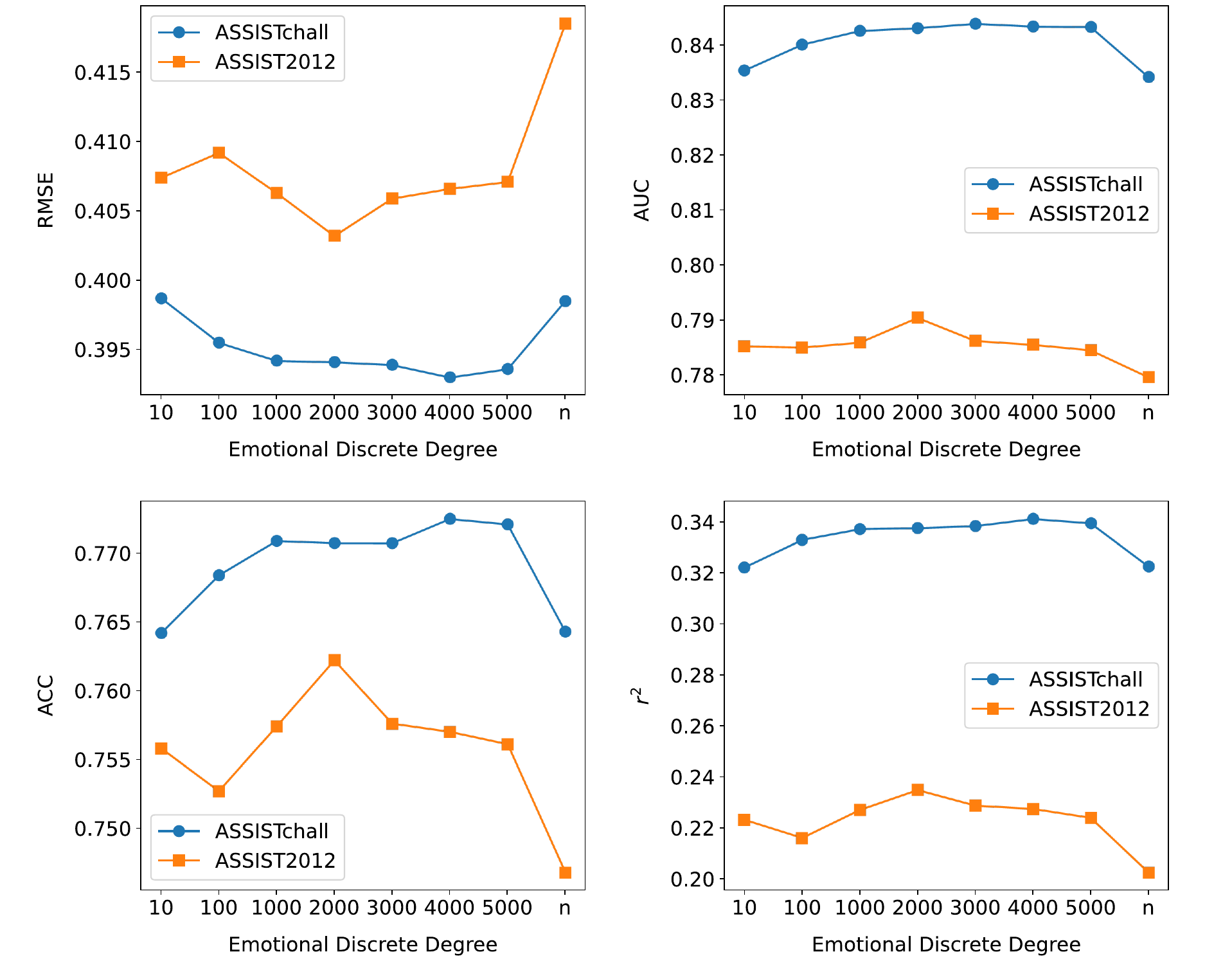}
\vspace{-0.3cm}
\caption{depicts the prediction results at different discrete levels (input dimension of Embedding), where n indicates that each emotion value corresponds to a category.}
\label{discretization}
\vspace{-0.3cm}
\end{figure}

\subsection{Analysis of Emotion-Boosted learning gains}
To illustrate the rationality and explainability of  the learning gains achieved under the influence of emotions,
we compare the knowledge state updating processes of the baseline LPKT and our DEKT. 
We analyze the variation in the mastery of knowledge concepts as a student sequentially completes 12 identical exercises involving 3 different concepts.
Figure \ref{Visualization}(a) shows the corresponding heatmap. There are several important observations in the figure.
Firstly, our DEKT method can capture more reasonable knowledge gains after interactions.
For example, in the first red rectangle, the knowledge gain of DEKT (0.74 to 0.79) is noticeably less than that of LPKT (0.71 to 0.85).
This is because, during interaction $t_5$, LPKT capture a basic knowledge gain by correctly answering the question.
However, DEKT notes the concentration at 0.2 and confusion at 0.6 during this interaction,  and perceives it as a low-quality interaction. 
Consequently, when computing the learning gains, a certain degree of negative constraint is applied to suppress substantial knowledge growth under negative emotions. 
The second red rectangle represents the opposite scenario.
Secondly, our DEKT method can also explain situations where the performance in answering questions does not align with the knowledge state.
For instance, in the orange rectangle, LPKT traces a high proficiency of 0.85 but fail to correctly solve the question.          LPKT can not explain this phenomenon, while DEKT can.
DEKT believes that due to the high confusion(0.6) of students, the subjective difficulty of the question increases.     
Simultaneously, their concentration(0.3) is low, leading students to easily overlook key information in the questions.     
As a result, students' knowledge state can not be adequately utilized, ultimately resulting in incorrect outcomes.
Fundamentally, the occurrence of this situation from an inadequate grasp of knowledge.   
Therefore, DEKT's 0.76 should be superior to LPKT's 0.85.

\subsection{Analysis of Personalized Emotional State}
In Figure \ref{Visualization}(a), while tracing changes in students' knowledge states using DEKT, we also record their personalized emotional states at these 12 moments.
Subsequently, we utilize t-SNE \cite{van2008visualizing} to reduce the 128-dimensional emotional states to a two-dimensional plane, as shown in Figure \ref{Visualization}(b).
Each circle represents the two-dimensional emotional state at the corresponding moment, with the size of the circle indicating the magnitude of a certain emotional representation value at the current moment.
Firstly, we analyze the relationship between personalized emotional state and response to questions:
(1) We observe that the positions of the circles within the red rectangle in Figure \ref{Visualization}(b) are very close, indicating minimal fluctuations in emotional states at these three moments.
This is because students grasp $c_{13}$ after only one incorrect response ($t_0$), indicating that students subjectively perceive this concept as relatively simple.  
Consequently, the establishment of emotional structure in students for solving this type of question is minimally affected.
As a result, in the subsequent three interactions where the student answer correctly ($t_1$, $t_2$, $t_3$), the emotional states exhibit minimal fluctuations. 
Furthermore, we also notice that during these three interactions, the improvement in knowledge state is not particularly significant (from 0.65 to 0.71 to 0.85).
(2) We also observe that at the four moments ($t_8$ to $t_{11}$), the emotional states are extremely unstable.
This is because the student answer the question about $c_{20}$ incorrectly twice (at $t_8$ and $t_9$).    
Even after answering correctly for the first time (at $t_{10}$), the improvement in knowledge state is not substantial.    
This indicates that mastering this concept is challenging.
The student lacks confidence in being able to answer correctly, thus each attempt at answering the question strongly influences their emotional state.
The above points also demonstrate that modeling emotional state is inseparable from knowledge state.
Secondly, we analyze the relationship between emotional state and emotional representations:
We have previously established that $c_{13}$ is relatively simple, whereas $c_{20}$ poses more difficulty.
From the graph, we can observe that when students continuously answer the simple concept $c_{13}$, their concentration gradually and slightly decreases, while their confusion remains relatively unchanged.  However, when continuously answering the difficult concept $c_{20}$, the changes in students' concentration and confusion are irregular and highly unstable.
This is also the theoretical basis for utilizing past habits of emotional representations to model personalized emotional state.

\subsection{Discrete level of emotion values}
The emotion values provided in the original dataset are not strictly continuous but highly precise discrete values.   
We consider that a specific emotion value does not carry meaningful information, and we are more interested in whether the emotion value is "high" or "low".  
Therefore, we discretize these finely detailed emotion values into more general categories.  
This simplifies the model, making it more interpretable and reduces the risk of overfitting.
To preserve the original information of the data as much as possible, the choice of the discrete scale (input dimension of the embedding layer) is particularly crucial.
As shown in Figure \ref{discretization}, we choose to discretize the emotion values into the following categories: 10, 100, 1000, 2000, 3000, 4000, 5000, n (where n indicates that each emotion value corresponds to a category).    
We observe that both excessively large and excessively small discrete level led to suboptimal results for the four evaluation metrics.
This is because when the discrete level is too small, such as splitting into 10 categories, values like 0.11 and 0.19 may be considered identical embeddings by the model, resulting in a significant loss of information.  
Conversely, when the discrete level is too large, treating each individual emotion value as a separate category could lead the model to learn irrelevant information, causing overfitting.  
For instance, emotion values such as 0.4882, 0.4883, and 0.4884 are practically indistinguishable, and if the model tries to learn distinct embeddings for them, it could adversely affect the results.
Hence, it is crucial to strike a balance in determining an appropriate discrete level that captures meaningful distinctions without introducing unnecessary complexity into the model.

\section{Conclusion}
This study aims to improve knowledge tracing techniques to more accurately trace students' knowledge states and proposes a novel Dual-State Personalized Knowledge Tracing with Emotional Incorporation (DEKT) model. 
The DEKT model first integrates students' emotional feedback to enhance learning gains after each interaction, thus achieving more accurate knowledge state. Secondly, it simulates the mechanism of student emotion generation and designs a method for predicting emotional representations based on personalized emotional state to achieve more reasonable predictions of student responses. 
Additionally, to enhance the model's generalization ability across different datasets, we utilize transfer learning techniques to design an enhanced version of DEKT, known as Transfer Learning-based Self-loop model (T-DEKT). 
Extensive experiments on three public datasets demonstrate that our approach not only outperforms existing knowledge tracing models but also exhibits excellent portability.
Therefore, this study brings new ideas and methods to the field of knowledge tracing, enhancing model accuracy and generalizability by considering student emotional feedback. 
This has significant implications for the future development of online learning systems, as it can provide more personalized and effective learning guidance for students.
In the future, we will explore other methods to improve the accuracy of emotion prediction.

\section*{Acknowledgments}
This work is supported by National Natural Science Foundation of China~(Grant No. 62106003, 62272435, U22A2094).

\bibliographystyle{IEEEtran}
\bibliography{mybibfile}

\begin{IEEEbiography}[{\includegraphics[width=1in, height=1.25in, clip, keepaspectratio]{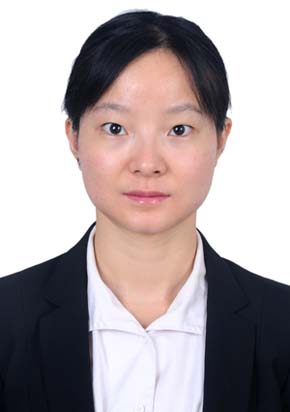}}]{Shanshan Wang}
received the B.E. degree, M.E. degree and the Ph.D. degree from Chongqing University in 2010, 2013
and 2020, respectively, in Chongqing, China. She is
currently a Associate Professor at the Institute of Physical Science and Information Technology, Anhui University, 
Hefei, China. Her research interests include computer vision, domain adaptation, and data mining.
\end{IEEEbiography}

\begin{IEEEbiography}[{\includegraphics[width=1in, height=1.25in, clip, keepaspectratio]{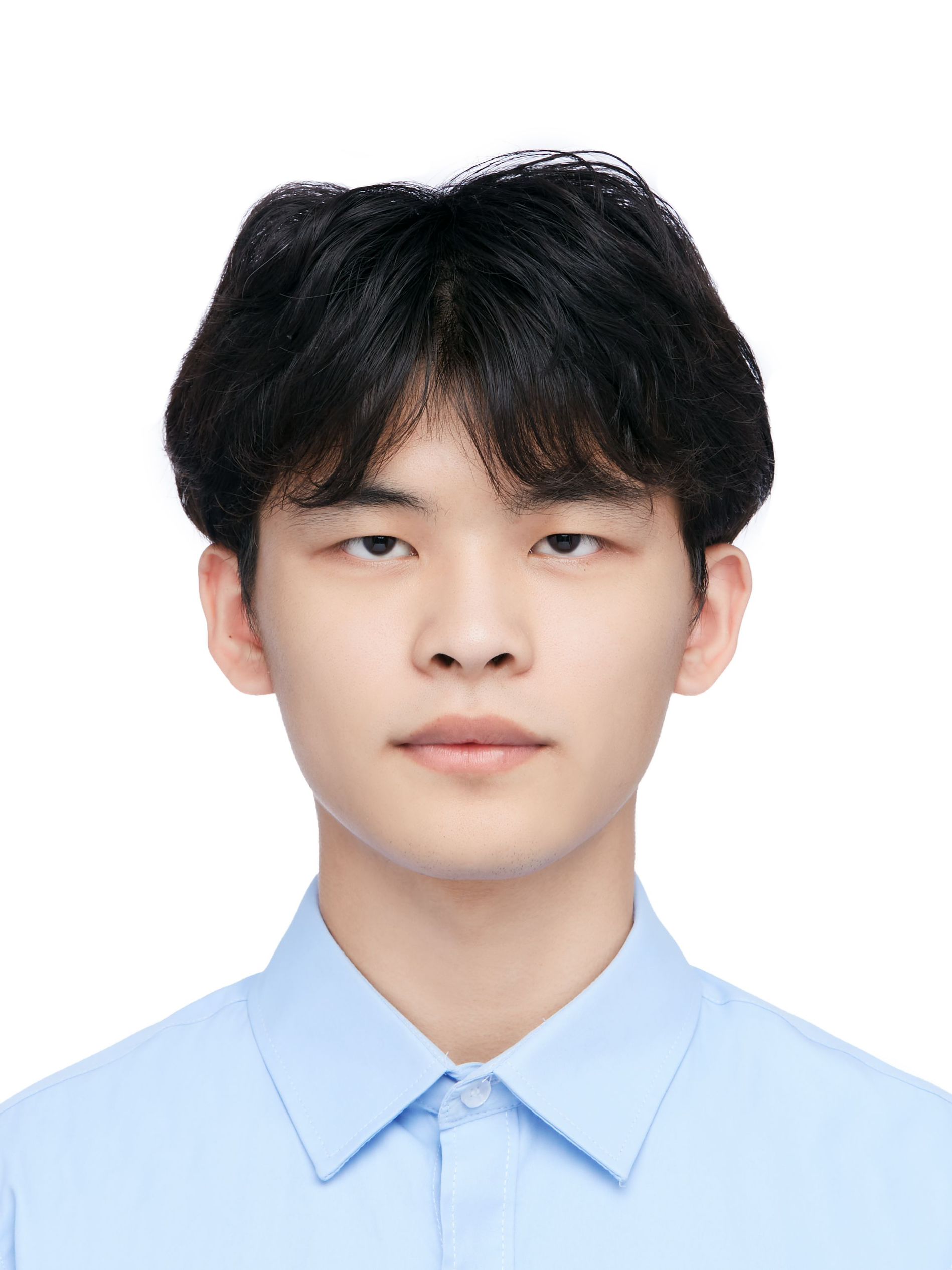}}]{Fangzheng Yuan}
received the B.S. degree in Computer Science and Technology from Shandong University of Science and Technology in 2022. He is currently pursuing the M.S. degree at the Institute of Physical Science and Information Technology, Anhui University.  His current research interests include data mining.
\end{IEEEbiography}

\begin{IEEEbiography}[{\includegraphics[width=1in, height=1.25in, clip, keepaspectratio]{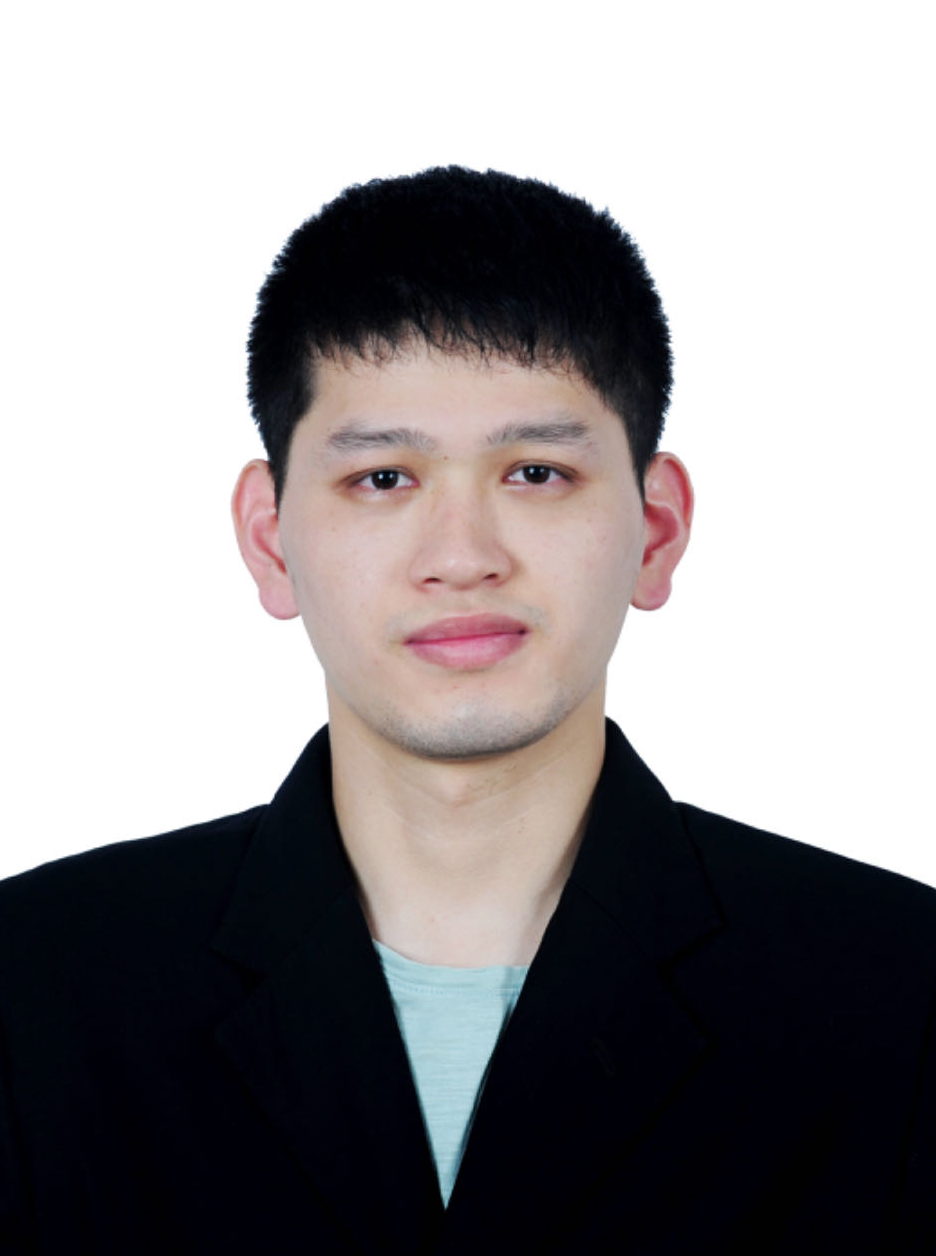}}]{Keyang Wang}
received his Master degree from Chongqing University, Chongqing, China in 2018.   He is currently an algorithm engineer at Zhejiang Dahua Technology Co., Ltd, Hangzhou, China.   His research interests include computer vision, machine learning and data mining.
\end{IEEEbiography}

\begin{IEEEbiography}[{\includegraphics[width=1in, height=1.25in, clip, keepaspectratio]{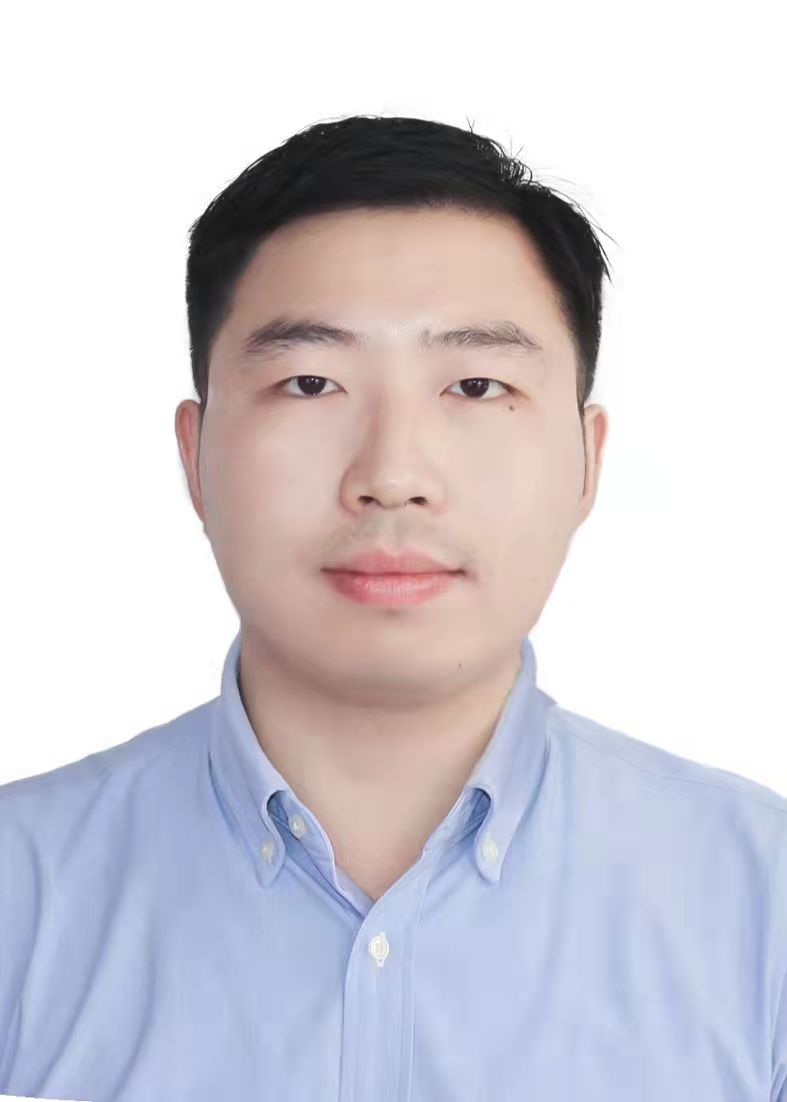}}]{Xun Yang}
received his Ph.D. degree from the Hefei University of Technology, Hefei, China, in 2017.  He is currently a Professor with the Department of Electronic Engineering and Information Science, University of Science and Technology of China (USTC). From 2015 to 2017, he visited the University of Technology Sydney (UTS), Australia as a joint Ph.D student. He was a Research Fellow with the NExT++ Research Center, National University of Singapore (NUS), from 2018 to 2021. His current research interests include information retrieval, cross-media analysis and reasoning, and computer vision. He regularly serves as the PC member and the invited reviewer for top-tier conferences and prestigious journals in multimedia and artificial intelligence, like the ACM Multimedia, IJCAI, AAAI, CVPR, and ICCV. He also served as a SPC member of the AAAI 2022 and an Area Chair of the ACM Multimedia 2022. He is an Associate Editor of the IEEE TRANSACTIONS ON BIG DATA (IEEE TBD).
\end{IEEEbiography}

\begin{IEEEbiography}[{\includegraphics[width=1in, height=1.25in, clip, keepaspectratio]{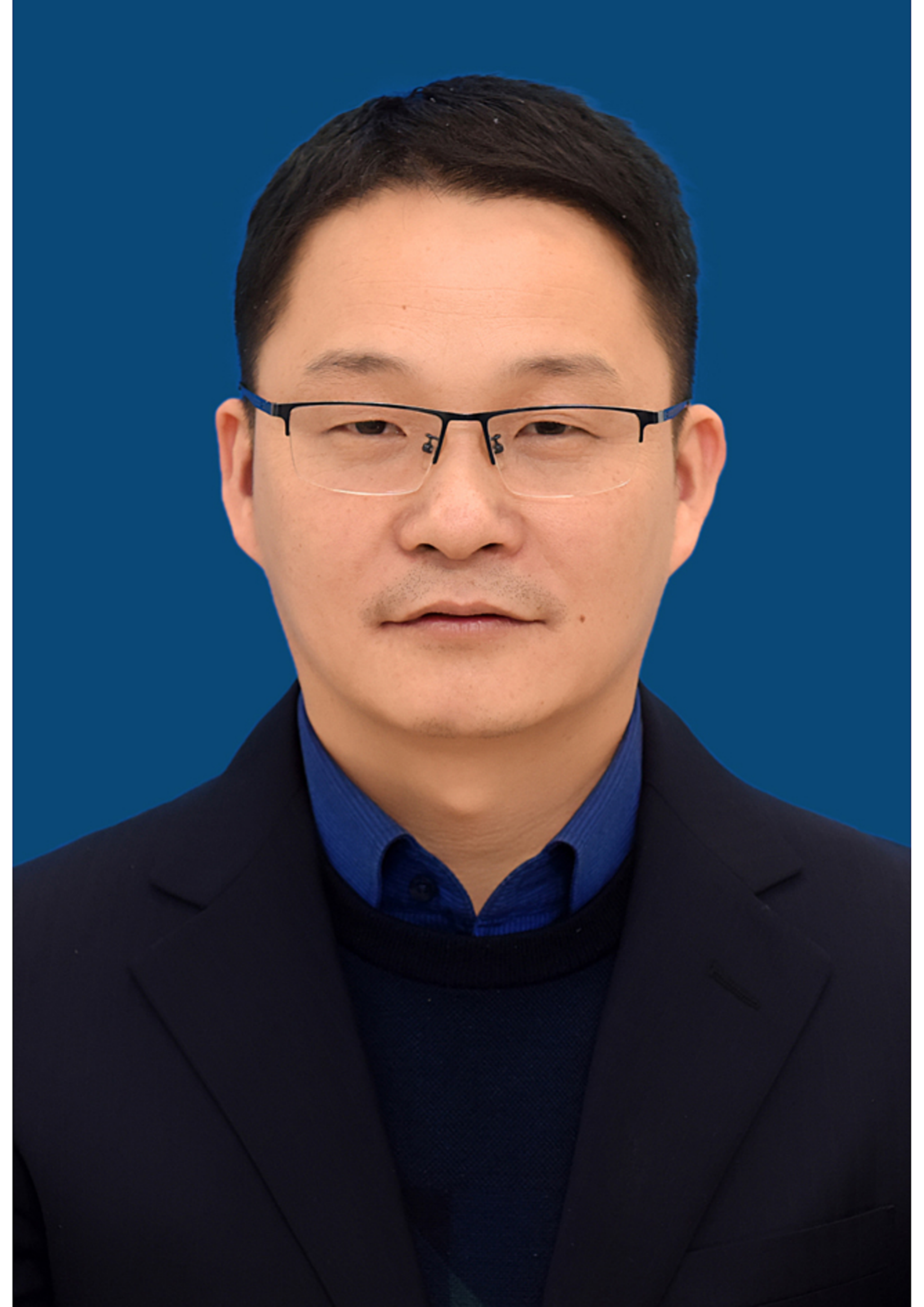}}]{Xingyi Zhang}
 (SM’18) received the B.Sc. degree
from Fuyang Normal College, Fuyang, China, 
in 2003, and the M.Sc. and Ph.D. degrees from
Huazhong University of Science and Technology, 
Wuhan, China, in 2006 and 2009, respectively.
He is currently a Professor with the School
of Computer Science and Technology, Anhui University, 
Hefei, China. His current research interests include
unconventional models and algorithms of
computation, evolutionary multi-objective optimization, 
and logistic scheduling. He is the recipient of the 2018
and 2021 IEEE Transactions on Evolutionary Computation Outstanding
Paper Award and the 2020 IEEE Computational Intelligence Magazine
Outstanding Paper Award.
\end{IEEEbiography}

\begin{IEEEbiography}[{\includegraphics[width=1in, height=1.25in, clip, keepaspectratio]{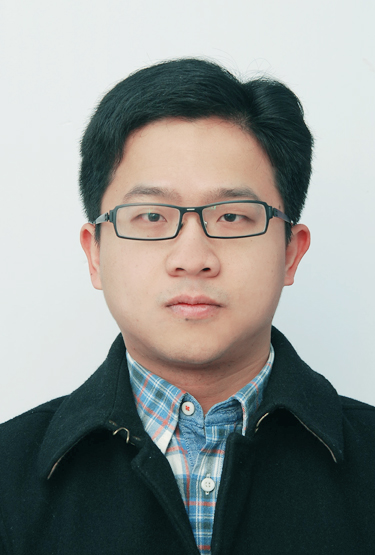}}]{Meng Wang} (Fellow, IEEE) received the B.E. and
Ph.D. degrees in the special class for the Gifted
Young and the Department of Electronic Engineering and Information Science, University of Science
and Technology of China (USTC), Hefei, China, 
in 2003 and 2008, respectively. He is currently a
Professor with the Hefei University of Technology, 
China. His current research interests include multimedia content analysis, computer vision, and pattern
recognition. He has authored more than 200 book
chapters and journal and conference papers in these
areas. He was a recipient of the ACM SIGMM Rising Star Award 2014.
He is an Associate Editor of the IEEE TRANSACTIONS ON KNOWLEDGE AND
DATA ENGINEERING (IEEE TKDE), IEEE TRANSACTIONS ON CIRCUITS
AND SYSTEMS FOR VIDEO TECHNOLOGY (IEEE TCSVT), IEEE TRANSACTIONS ON MULTIMEDIA (IEEE TMM), and IEEE TRANSACTIONS ON
NEURAL NETWORKS AND LEARNING SYSTEMS (IEEE TNNLS).
\end{IEEEbiography}

\vspace{11pt}

\vfill

\end{document}